\documentclass{article}

\usepackage{arxiv}

\usepackage[utf8]{inputenc} 
\usepackage[T1]{fontenc}    
\usepackage{hyperref}       
\usepackage{url}            
\usepackage{booktabs}       
\usepackage{amsfonts}       
\usepackage{nicefrac}       
\usepackage{microtype}      
\usepackage{lipsum}		
\usepackage{graphicx}
\usepackage[numbers]{natbib}
\usepackage{doi}

\usepackage{subcaption}
\usepackage{amsmath,amssymb,amsfonts}
\usepackage{adjustbox}
\usepackage{array}

\newcommand{\bbR}{{\mathbb R}}
\newcommand{\SO}{\mathcal{SO}}

\title{RIOT: Recursive Inertial Odometry Transformer for Localisation from Low-Cost IMU Measurements}

\author{ \href{https://orcid.org/0000-0001-7912-8741}{\includegraphics[scale=0.06]{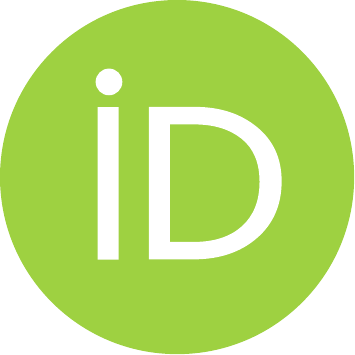}\hspace{1mm}James Brotchie}\thanks{Corresponding author.} \\
	Department of Science\\
	RMIT University\\
	Melbourne, VIC 3000, Australia \\
	\texttt{s3238455@student.rmit.edu.au} \\
	\And
	Wenchao Li \\
	Department of Science\\
	RMIT University\\
	Melbourne, VIC 3000, Australia \\
	\texttt{wenchao.li@rmit.edu.au} \\
	\And
	Andrew D.~Greentree \\
	Department of Science\\
	RMIT University\\
	Melbourne, VIC 3000, Australia \\
	\texttt{andrew.greentree@rmit.edu.au} \\
	\And
	Allison Kealy \\
	Victorian Department of Environment, Land, Water and Planning\\
	RMIT University\\
	Melbourne, VIC 3000, Australia \\
	\texttt{allison.kealy@rmit.edu.au} \\
}

\renewcommand{\shorttitle}{\textit{arXiv} Template}

\hypersetup{
pdftitle={RIOT: Recursive Inertial Odometry Transformer for Localisation from Low-Cost IMU Measurements},
pdfsubject={q-bio.NC, q-bio.QM},
pdfauthor={David S.~Hippocampus, Elias D.~Striatum},
pdfkeywords={First keyword, Second keyword, More},
}

\begin{document}
\maketitle

\begin{abstract}
	Inertial localisation is an important technique as it enables ego-motion estimation in conditions where external observers are unavailable. However, low-cost inertial sensors are inherently corrupted by bias and noise, which lead to unbound errors, making straight integration for position intractable. Traditional mathematical approaches are reliant on prior system knowledge, geometric theories and are constrained by predefined dynamics. Recent advances in deep learning, that benefit from ever-increasing volumes of data and computational power, allow for data driven solutions that offer more comprehensive understanding. Existing deep inertial odometry solutions rely on estimating the latent states, such as velocity, or are dependant on fixed sensor positions and periodic motion patterns. In this work we propose taking the traditional state estimation recursive methodology and applying it in the deep learning domain. Our approach, which incorporates the true position priors in the training process, is trained on inertial measurements and ground truth displacement data, allowing recursion and to learn both motion characteristics and systemic error bias and drift. We present two end-to-end frameworks for pose invariant deep inertial odometry that utilise self-attention to capture both spatial features and long-range dependencies in inertial data. We evaluate our approaches against a custom 2-layer Gated Recurrent Unit, trained in the same manner on the same data, and tested each approach on a number of different users, devices and activities. Each network had a sequence length weighted relative trajectory error mean $\leq0.4594$m, highlighting the effectiveness of our learning process used in the development of the models.
\end{abstract}


\section{Introduction}
Inertial odometry is crucial in mobile agents as it facilitates ego-motion in many applications such as autonomous driving, health/activity
monitoring, indoor navigation, human-robot interaction and augmented/virtual reality. Inertial measurement units (IMUs) are low-power, offer high privacy, and are robust in various environments. As such, offer a cheap and completely ego-centric means of localisation. IMUs typically consist of a 3D accelerometer, 3D gyroscope and 3D magnetometer. By accurately integrating data from IMUs and other sensors, it is possible to build a reliable system for estimating the motion and position for autonomous systems and pedestrian navigation. However, low-cost inertial sensors have high levels of noise and biases, causing unbound system error growth in long term inertial navigation \cite{el2007analysis}. 

Neural networks have the ability to employ continuous activation functions that inherently understand time and are capable of modelling complex non-linear system behaviours, which are typically too complex for classical mathematical approaches \cite{NNapplications}. Recurrent neural networks (RNNs) have long been the primary choice for sequence-to-sequence modelling. Most existing deep inertial navigation solutions utilise RNNs, some supplementing with convolutional neural networks (CNNs) (see Section~\ref{rw}). However, these architectures have well documented limitations such as an inability to capture long-term dependencies and sequential computation that cannot be parallelised \cite{pascanu2013difficulty}. 

These deficiencies lead to the development of new architectures. The most notable being self-attention based Transformer models, first proposed in \cite{AIAYN}, which since inception has become ubiquitous in natural language processing (NLP). The success seen in NLP has proliferated their use in a number of domains. Recently we have seen Transformers employed in computer vision (CV) \cite{khan2022transformers}, NLP \cite{bert}, time-series forecasting \cite{tsTransformer}, image recognition/production \cite{image}, text summarisation \cite{generating}, speech recognition \cite{asr} and music generation \cite{music}. These implementations have displayed the networks ability to model long dependencies between input sequence elements and can be processed in parallel, contrasting RNNs. As noted in \cite{merkx2020comparing}, these capabilities have the advantageous property of resolving the memory bottleneck commonly found in RNNs. Additionally, a comparison in the effectiveness of a long short term memory (LSTM) network (a RNN variant) and self-attention based Transformer showed significant performance gains in self-attention based techniques on data sets with long range dependencies \cite{radford2018improving}. 

In contrast to CNNs, Transformers do not necessitate design specifications and are proficient in handling set functions. Additionally, their uncomplicated architecture facilitates the processing of diverse modalities through the utilisation of homogeneous processing units, which have proven to exhibit remarkable scalability to both large networks and datasets. The incorporation of self-attention mechanism in neural networks enable the inputs to engage with one other and to be evaluated according to their correlation with the final prediction. Despite extensive investigation into this formulation, limited research has been conducted utilising self-attention and unprocessed sequential readings obtained from low-cost, noisy inertial sensors in the inertial odometry domain. The substantial achievements achieved in other sequence-to-sequence learning problems indicate that the application of self-attention based techniques could eliminate the need for accurate dynamic models and offer a more robust solution compared to previous RNN or CNN based methodologies.

\section{Related Work}\label{rw}
Recent work has shown that the implementation of an accurate inertial odometry algorithm can serve as a foundation for a more robust and reliable ego-motion estimation system through the fusion of inertial sensors. The following details previous deep inertial odometry proposals in the literature: 

Authors of \cite{yan2018ridi} demonstrated the feasibility of using a smartphone's 6D inertial sensor (consisting of 3D accelerations and 3D angular velocities) and orientation estimation provided by its application programming interface (API) to locate pedestrians. Termed RIDI, their approach leverages patterns in natural human movement to learn how to predict velocity and correct linear accelerations using linear least squares. Recent advances in deep learning (DL) has further accelerated data driven based inertial navigation. In \cite{asraf2021pdrnet}, a deep learning approach called PDRNet was developed for pedestrian dead reckoning (PDR). PDRNet consists of a classification network for smartphone location recognition and a regression network for determining the change in distance and heading. These approaches rely heavily on a large number of carefully tuned parameters and user's walking habits, which leaves them fundamentally susceptible to rapid drift and are not generalisable.  IONet \cite{chen2018ionet} utilise a bidirectional long short-term memory (Bi-LSTM) and kinetic models to regress the magnitude velocity and the changing rate of direction. RONIN \cite{yan2019ronin} take ResNet \cite{he2016deep}, a LSTM built for CV, as a backbone to again regress a velocity vector. Authors of \cite{khorrambakht2021deep} apply preintegration and an LSTM as a solution to supplement the IMU motion model for deep inertial odometry. These unified deep neural networks provide more robust solutions in highly dynamic conditions, however are still reliant on direct integration. Additionally, these methods rely on IMU orientation, are all limited in dimensionality and dependent on dynamical models that require prior knowledge in the dynamics of the system. 

In \cite{liu2020tlio} the authors present TLIO, again using ResNet, to regress 3D displacement estimates and the uncertainty, allowing them to tightly fuse the relative state measurement into a stochastic cloning extended Kalman filter (EKF) to solve for pose, velocity and sensor biases. Owing to its reliance on an EKF it was shown to be susceptible to system failure during highly dynamic or unusual motion, which is in line with previous work \cite{brotchie2021evaluating}. Similarly, in \cite{sun2021idol}, the authors propose a hybridised approach using an LSTM and EKF in a modular design that consists of orientation and position subsystems, termed IDOL. Having a dedicated orientation module that included 3D magnetometer readings proved advantageous in contrast to previous approaches that are reliant on the system API. Authors of \cite{wang2021pose} present a novel loss formulation for smartphone-based deep inertial odometry. The authors use a ResNet-style neural network utilising 2-second inertial signals from an IMU to estimate the average velocity and direction of movement. It's noted in this work that despite the obvious benefits of incorporating the magnetometer readings, the network would not converge. 

The most recent approach is given in \cite{cao2022rio} and iterated in \cite{wang2022a2dio}, where the authors propose attention driven rotation-equivariance-supervised inertial odometry, based off 6D IMU readings. They adopted ResNet to show that adding a self-supervised auxiliary task based on rotation-equivariance can improve the performance of the model when it is jointly trained and can be further improved with a Test-Time Training strategy. In their follow up, the authors propose a hybrid neural network model for inertial odometry that combines a CNN block with attention mechanisms and a Bi-LSTM network. The CNN block is used to extract spatial features from normalised 6D IMU measurements, and the attention mechanisms, which include a spatial attention mechanism and a channel attention mechanism, are used to refine these features. The Bi-LSTM network is then used to capture temporal features. 

The effectiveness of data-driven solutions in inertial odometry is well-documented, but these approaches share a common issue in network design. A well-designed network can improve performance in various applications \cite{abiodun2018state} and IMU data, collected at high frequencies, can be challenging to process using traditional machine learning approaches such as RNNs and CNNs. One drawback of using RNNs for IMU data processing is the issue of "washout", whereby the network's ability to remember past inputs diminishes over time \cite{mohajerin2019multistep}, making it difficult to accurately process long sequences of data. On the other hand, CNNs require deep architectures to cover large enough receptive fields to effectively process IMU data, which can result in significant computational expenses during training and deployment \cite{alzubaidi2021review}.

\section{Preliminaries}
In this work we propose a new approach to deep inertial odometry using self-attention encoder-decoder based network blocks. To mitigate the challenges associated with high frequency time series data we propose relying entirely on the self-attention mechanism to compute representations of the inputs and outputs, rather than using sequence-aligned RNNs or convolutions. Given that the self-attention mechanism is poised to be the primary means of information extraction from input to output generation, it holds the promise of greater efficiency and flexibility compared to relying on RNNs or CNNs. This is because self-attention mechanisms have the ability to capture long-range dependencies in data, while also allowing for parallelisation during the training phase. Additionally, self-attention mechanisms provide a degree of interpretability by allowing the model to identify the most important input features at each time step. 

We term our approaches: Recursive Inertial Odometry Transformer (RIOT) and  Attitude Recursive Inertial Odometry Transformer (ARIOT). To the best of our knowledge, our approaches are the only networks that leverage self-attention and all available IMU information (from a $3$D accelerometer, $3$D gyroscope and $3$D magnetometer) to provide an end-to-end, $3$D inertial odometry solution. 

The ARIOT model is a hierarchical transformer, and differs from RIOT by incorporating of an additional, internal attitude estimation network that regresses the orientation of the IMU from the sensor measurements. This subsystem benefits from the self-attention based framework design in \cite{brotchie2022leveraging}.The output is used in the odometry network to further regress the accelerometer readings and prior position to give updated IMU localisation. The effectiveness of RIOT and ARIOT is validated on unseen sequences in their entirety from different users, activities and smart phone IMU devices. 

\section{Problem Formulation}
\subsection{Sensor Models}\label{sec:sensors}
First we consider the problem of modelling measurements from a $9$D IMU. It is implicit that these systems are characterised by high noise levels and time-varying additive biases. The available measurements from a typical IMU are from three-axis rate gyros, three-axis accelerometers and three-axis magnetometers. The reference frame of the IMU is termed the body frame ($B$), which is rotated with respect to some fixed inertial frame ($I$), e.g., the Earth-centered inertial (ECI) frame or the North-East-Down (NED) frame. However, for brevity these reference frames are assumed and not incorporated into the notation.

The gyroscope measures the angular velocity of $B$ relative to $I$, corrupted by a slowly varying bias and noise. Therefore, we can define the gyroscope measurements, $\mathbf{I}_{\omega, t}$, as
\begin{equation}\label{gyr}
\mathbf{I}_{\omega, t}=\boldsymbol{\omega}_{t}+\boldsymbol{\delta}_{\omega, t}+\boldsymbol{e}_{\omega, t} 
\end{equation}
where $\boldsymbol{\omega}_{t}$ is the true angular velocity at each time instance $t$, $\boldsymbol{\delta}_{\omega, t}$ denotes the time-varying bias and $\boldsymbol{e}_{\omega, t}$ is the noise, typically assumed to be Gaussian, $\boldsymbol{e}_{\omega, t} \sim \mathcal{N}\left(0, \Sigma_\omega\right)$.

The accelerometer measures the linear acceleration of $B$ relative to $I$. Again with added noise and bias, the accelerometer measurements, $\mathbf{I}_{a, t}$, are given by
\begin{equation}\label{acc}
\mathbf{I}_{a, t}=\mathbf{f}_t+\boldsymbol{\delta}_{a, t}+\boldsymbol{e}_{a, t}
\end{equation}
where $\mathbf{f}_t$ is the specific force at each time instance $t$, and $\boldsymbol{\delta}_{a, t}$ and $\boldsymbol{e}_{a, t}$ denote the bias and noise, respectively, with $\boldsymbol{e}_{a, t} \sim \mathcal{N}\left(0, \Sigma_\omega\right)$.

Magnetometers provide information about the direction and intensity of the local magnetic field surrounding the sensor. The local magnetic field is composed of the Earth's magnetic field as well as any additional magnetic fields that arise due to the existence of magnetic materials. As magnetometer measurements are used primarily in attitude determination, we assume the magnitude of the local magnetic field vector, denoted $\mathbf{m}^l$, is equal to 1 --  i.e. $\left\| \mathbf{m}^l \right\|=1$. Assuming that the magnetometer only measures the local magnetic direction, its measurements, $\mathbf{I}_{\mathrm{m}, t}$, can be modelled as
\begin{equation}\label{mag}
\mathbf{I}_{m, t}=\mathbf{R}_t^{bn} \mathbf{m}^l +\boldsymbol{e}_{m, t},
\end{equation}
where $\mathbf{R}_t^{bn}$ denotes the rotation matrix from navigation to body frame and $\boldsymbol{e}_{\mathrm{m}, t} \sim \mathcal{N}\left(0, \Sigma_{\mathrm{m}}\right)$ is the Gaussian noise. 

By incorporating magnetometer measurements we enable the system to determine its initial attitude. This is predicated on the principle that, give a set of two or more linearly independent vectors in two distinct reference frames, the rotation between said frames can be calculated. The underlying assumption here is that the accelerometer only measures the gravity vector and the magnetometer only measures the local magnetic field. Hence we have four linearly independent vectors: measurements $\mathbf{I}_{a, t}$ and $\mathbf{I}_{m, t}$, the local gravity vector $\boldsymbol{g}^n$, and the local magnetic vector, $\mathbf{m}^l$. Whilst this is seen as a major advantage, it does come with the drawback of requiring local magnetic field knowledge in order to transform Eq.~\ref{mag} into local coordinates.

\subsection{Attitude and Position Estimation}
Traditional attitude estimation approaches rely on gyro integration as the baseline for deriving the attitude. However, it is well documented that gyroscope measurements lack the information to give absolute attitude determination. Therefore, applying numerical gyro integration results in an accumulated error that grows boundlessly. As such, specific force measurements from an accelerometer are often used in tandem with magnetic field measurements from a magnetometer to correct the estimate, as they provide information to the absolute angular position. These methods typically involve complex mathematical models and computations, and require prior state knowledge and specific sensor parameters \cite{crassidis2007survey}.

Analogous to attitude estimation, traditional methods for position estimation are also susceptible to unbounded errors due to the lack of information for absolute position change. To overcome this limitation, we propose using self-attention and raw data in gradient descent optimisation to analyse and retain information related to accelerometer error and bias over long sequences. The relevant features are extracted by the attention mechanism to learn the relationship between acceleration, attitude, and position. In the case of ARIOT, this method also have the caveat of recognising and compensating for attitude estimation errors in the initial attitude estimation network, seen in \cite{brotchie2022leveraging}.

We use the accelerometer and gyroscope measurements as inputs to the dynamics for the purpose of estimating the position. The state vector includes the position and a quaternion parametrisation of the attitude (detailed in Section~\ref{sec:ariot}). We use the inertial measurements along with prior positions to estimate the attitude and position.

The dynamics of the position for an interval of time $\Delta t$ are given by the equation
\begin{equation}
\mathbf{p}_{t+1}=\mathbf{p}_t+\Delta t \boldsymbol{v}_t+\frac{\Delta t^2}{2}\left(\mathbf{R}_t^{nb}\left(\mathbf{I}_{a, t}-\boldsymbol{\delta}_{a, t}\right)+\boldsymbol{g}^n+\boldsymbol{e}_{a, t}\right) \\
\end{equation}
where $\boldsymbol{v}_t$ denotes the velocity and time$t$ and $\mathbf{R}_t^{nb}$ is the rotation matrix from body to navigation frame. We switched the sign on the noise term for convenience. The dynamics of the attitude is then given by

\begin{equation}
\mathbf{q}_{t+1}=\mathbf{q}_t \odot \exp _q\left(\frac{\Delta t}{2}\left(\boldsymbol{I}_{\omega, t}-\boldsymbol{\delta}_{\omega, t}-\boldsymbol{e}_{\omega, t}\right)\right) \odot \exp_q\left(\frac{\Delta t}{2}f\left(\boldsymbol{I}_{a, t}, \boldsymbol{I}_{m, t}\right)\right)
\end{equation}
where $f( \cdot )$ is a function of the accelerometer and magnetometer measurements to calculate the correction term for the quaternion and the notation $\odot$ denotes the quaternion multiplication given by

\begin{equation}
\begin{pmatrix}j_{1}\\ j_{2}\\ j_{3}\\ j_{4}\end{pmatrix} \odot \begin{pmatrix}k_{1}\\ k_{2}\\ k_{3}\\ k_{4}\end{pmatrix} = \begin{pmatrix}j_{1}k_{1} - j_{2}k_{2} - j_{3}k_{3} - j_{4}k_{4}\\ j_{1}k_{2} + j_{2}k_{1} + j_{3}k_{4} - j_{4}k_{3}\\ j_{1}k_{3} - j_{2}k_{4} + j_{3}k_{1} + j_{4}k_{2}\\ j_{1}k_{4} + j_{2}k_{3} - j_{3}k_{2} + j_{4}k_{1}\end{pmatrix}.
\end{equation}

A significant advantage of traditional state estimation algorithms over neural networks is the retention of prior state estimate to update subsequent states. This allows for the algorithm to use past information to correct or refine the current estimate, which improves accuracy and reliability. In contrast, neural networks are typically trained to make predictions based on input data without explicit retention of past estimates. 

We recognise that RNNs are somewhat the exception here as they can be designed to have internal state memory that can retain past state information, however this doesn't actually give the network the desired recursive property. Instead it acts more as a pseudo recursion in which the hidden states store some information from previous steps and use it to influence future steps. This distinction is crucial in understanding the nature and limitations of RNNs in terms of recursive behaviour. Furthermore, networks that adopt this come with the aforementioned memory bottlenecks and vanishing gradient drawbacks. 

\section{Proposed Solution}
\subsection{Network Components}\label{network}
An overview of the models is depicted in Figure.~\ref{ariot} and Figure.~\ref{riot}. Here we will introduce the common components found in both networks. The modular specific adaptations will follow. In model design we follow the original NLP Transformer proposed in \cite{AIAYN}, comprising of encoder-decoder blocks and multi-head attention (MHA). An advantage of this intentionally straightforward system design is that its efficient to implement and provides an out of the box solution. The input of the standard Transformer is a 1D sequence of token embeddings. To handle IMU data, the sequence embeddings are expanded to $N$-dimensions corresponding to feature inputs, each with a set of additional position embeddings, which represent the temporal information. The network produces a sequence of representations for each input time-step, which is then used as feature vectors in downstream tasks. 

\subsubsection{Positional Encoding}
In the Transformer model, as described in \cite{AIAYN}, relative sequential position is not explicitly encoded. To incorporate relative sequential position information, we add sinusoidal position encoding functions over the inputs before the first layer. The values of the encoding are calculated using the trigonometric functions $\sin$ and $\cos$. The argument of these functions is the product of the sequence position $(pos)$ and a scaling factor $10000^{2i / d_{\text {model }}}$, where $i$ is an index variable, $0 \le i \le \frac{d_{\text {model }} - 1}{2}$, used to calculate different dimensions of the positional encoding vector. For each value of $i$, two dimensions of the positional encoding vector are calculated, one using the sine function ($PE_{(pos, 2i)}$) and one using the cosine function ($PE_{(pos, 2i + 1)}$). The idea behind using both the sine and cosine functions is capture both the magnitude and phase of the sequential position information i.e. relative order and distance between elements in the sequence. This in turn allows the model to attend different parts of input sequence at different stages of processing. Additionally, this approach allows the model to generalise to different sequence lengths and attend to elements based on their relative positions rather than their absolute positions. This can be useful when the model needs to handle sequences of different lengths or in cases of mismatched sampling \cite{coviello2020study}. The positional encoding in both networks are defined by

\begin{equation}\label{encoding}
\begin{aligned}
P E_{(p o s, 2 i)} &=\sin \left(p o s / 10000^{2 i / d_{\text {model }}}\right) \\
P E_{(p o s, 2 i+1)} &=\cos \left(p o s / 10000^{2 i / d_{\text {model }}}\right)
\end{aligned}
\end{equation}
where $pos$ denotes the position, $i$ the dimension and $d_{\text {model }}$ is the model dimensionality; in this work $d_{\text {model }}$ is $64$ and $224$ for the attitude and position networks (see Section~\ref{ablations}), respectively. 

\subsubsection{Self-Attention}
Self-attention sublayers in these networks employ $h=2$ heads. The self-attention mechanism works by first projecting the IMU measurements into a higher-dimensional space using a linear transformation parameterised by a set of weights $\mathbf{W}^Q, \mathbf{W}^K, \mathbf{W}^V \in \mathbb{R}^{d_I \times d_{\text {model }}}$. This projection is parameterised by a set of weights, which are learned during training. These parameter matrices are unique per layer and attention head. The transformed input data is then passed through a function (often called the "attention function") which produces a set of attention weights for each input element, representing the importance of each input element in regressing to an attitude or position estimation. 

Each attention head operates on an input sequence $\mathbf{I} = (I_1, \dots, I_n)$ of $n$ elements where $I_i \in \bbR^{d_I}$. A new sequence of the same length in computed as $\mathbf{z} = (z_1, \dots, z_n)$, where $z_i \in \bbR^{d_{\text {model }}}$, and each output element is computed as a weighted sum of a linearly transformed inputs per
\begin{equation}
z_i = \sum_{j=1}^T \alpha_{i,j} \left( x_j \mathbf{W}^V\right)
\end{equation}
where each weight coefficient, $\alpha_{i,j}$ is computed through a softmax function, which normalises the compatibility scores for each element to produce a probability distribution over the input sequence 

\begin{equation}
\alpha_{i,j} = \frac{\exp e_{i,j}}{\sum_{k=1}^n \exp e_{i,k}} 
\end{equation}
$e_{i j}$ is computed using a compatibility function that compares two input elements,
\begin{equation}
e_{i j}=\frac{\left(x_i \mathbf{W}^Q\right)\left(x_j \mathbf{W}^K\right)^T}{\sqrt{d_z}}
\end{equation}
Scaled dot product is used as the compatibility function to enable efficient computation. Linear transformations of the inputs add sufficient expressive power. The self-attention layer is implemented using MHA. 

In the context of using IMU information to estimate an attitude quaternion or position, the self-attention mechanism is used to weight the different sensor measurements differently, depending on how relevant they are to the position estimate. For example, the gyroscope measurements may be given a higher weight than the accelerometer measurements when estimating rotational motion, while the accelerometer measurements may be given a higher weight when estimating linear acceleration. 

\subsubsection{Encoder} 
The element-wise addition of the input vector and positional encoding vector is fed into two identical encoder layers. Each encoding layer is made up of two sub-layers: a MHA sub-layer and a fully connected feed-forward (FF) sub-layer. In the case of the ARIOT attitude module, we trialled a number of convolution layers to extract the spatial structure features of the data, however the self-attention mechanism proved enough to capture the relevant information and no benefit was seen.

Our encoder follows the Query--Key--Value model, proposed in \cite{AIAYN}, where the scaled dot-product attention used is given by
\begin{equation}\label{eq:softmax}
\operatorname{Attention}(\mathbf{Q}, \mathbf{K}, \mathbf{V})=\operatorname{softmax}\left(\frac{\mathbf{Q} \mathbf{K}^{{T}}}{\sqrt{D_{k}}}\right) \mathbf{V}
\end{equation}
where the input, $\mathbf{I}$, is used to obtain the queries $\mathbf{Q} = \mathbf{I}(k)\mathbf{W}^Q \in \mathbb{R}^{N \times D_{k}}$, keys $\mathbf{K} = \mathbf{I}(k)\mathbf{W}^K \in \mathbb{R}^{M \times D_{k}}$ and values \linebreak $\mathbf{V} = \mathbf{I}(k)\mathbf{W}^V \in \mathbb{R}^{M \times D_{v}}$; each $\mathbf{W}$ is the respective weight matrices updated during training, and $N, M$ denote the lengths of queries and keys (or values) and $D_{k}, D_{v}$ denote the dimensions of keys (or queries) and values. The MHA consists of $H$ different sets of learned projections instead of a single attention function as

\begin{equation}
\text{MultiHeadAttn}(\mathbf{Q}, \mathbf{K}, \mathbf{V})= \text{Concat}(\text{head}_{1}, \dots, \text{head}_{H}) \mathbf{W}^{O}\notag
\end{equation}
 where $\text{ head}_{i}=$ Attention $\left(\mathbf{Q} \mathbf{W}_{i}^{Q}, \mathbf{K} \mathbf{W}_{i}^{K}, \mathbf{V}_{{i}}^{V}\right)$. 
The projections are parameter matrices $W_{i}^{Q} \in \mathbb{R}^{D_{\text {model }} \times D_{k}}, W_{i}^{K} \in \mathbb{R}^{D_{\text {model }} \times D_{k}}, W_{i}^{V} \in \mathbb{R}^{D_{\text {model }} \times D_{v}}$ and $W^{O} \in \mathbb{R}^{h D_{v} \times D_{\text {model }}}$.
In this work, we employ $h=2$ parallel attention layers, or heads. For each, we use \linebreak $D_{k}=D_{v}=D_{\text {model }}/h$. 

In addition to the attention sub-layers, each encoder/decoder layer consists of a fully connected FF network, consisting of linear transformation and activation functions. We use a $\text{LeakyRe}LU$ \cite{leakyRelu} activation in the FF network as follows  

\begin{equation}
\text{LeakyRe}LU(x)= \begin{cases}x, & \text { if } x \geq 0 \\ 1 \times 10^{-3} \cdot x, & \text { otherwise }\end{cases}
\end{equation}

The point-wise FF network is a fully connected module
\begin{equation}
\text{FFN}\left(\mathbf{H}^{\prime}\right)=\text{LeakyRe}LU\left(\mathbf{H}^{\prime} \mathbf{W}^{1}+\mathbf{b}^{1}\right) \mathbf{W}^{2}+\mathbf{b}^{2}
\end{equation}
where $\mathbf{H}^{\prime}$ is the output of the previous layer, $\mathbf{W}^{1} \in \mathbb{R}^{D_{m} \times D_{f}}$, $\mathbf{W}^{2} \in \mathbb{R}^{D_{f} \times D_{m}}, \mathbf{b}^{1} \in \mathbb{R}^{D_{f}}$and $\mathbf{b}^{2} \in \mathbb{R}^{D_{m}}$ are trainable parameters, and $D_f$ denotes the inner-layer dimensionality. Each sub-layer has a Layer Normalisation Module inserted around each module. That is,
\begin{equation}
\mathbf{H}^{\prime} =\text{LayerNorm}\left(\text{SelfAttn}(\mathbf{X})+\mathbf{X}\right) \\
\end{equation}
where $\text{SelfAttn}(\cdot)$ denotes self-attention module and $\text{LayerNorm}(\cdot)$ the layer normal operation. The resultant vector is then fed into the decoder. 

\subsubsection{Decoder} 
The decoder is composed of 2 identical layers. The decoder contains the sub-layers found in the encoder, with the addition of a third sub-layer that performs MHA over the output vector from the encoder. The MHA mechanism allows the model to attend to multiple parts of the input sequence in parallel, allowing it to capture a more detailed and nuanced representation of the input. This is achieved by dividing the attention mechanism into multiple "heads". Each head performs attention with a different linear projection. Additionally, the self-attention mechanism in the decoder stacks prevent positions from influencing subsequent positions to ensure that predictions at $k$ can depend only on the known outputs at or before ${k-1}$. In our attitude network, the output maps the final layer into the estimated quaternion through a hyperbolic tangent. For the position, no outbound function is used past linearisation.

\subsection{Attitude Recursive Inertial Odometry Transformer}
\label{sec:ariot}
The Attitude Recursive Inertial Odometry Transformer is a hierarchical framework composing of 2 self-attention based encoder-decoder networks. The foundation for the initial network is based on previous work \cite{brotchie2022leveraging} and functions to regress attitude estimation from $9$D inertial measurements (from Eq.~\ref{gyr}, Eq.~\ref{acc} and Eq.~\ref{mag}). This allows for the componential estimation of both attitude and position estimation in a single framework, providing a robust solution for inertial odometry. The use of self-attention mechanisms within both modules allow for the modelling of long-term dependencies in the data, effectively handling the high dynamic motion present over long sequences. 

\begin{figure*}
\centering
\includegraphics[width=0.99\textwidth]{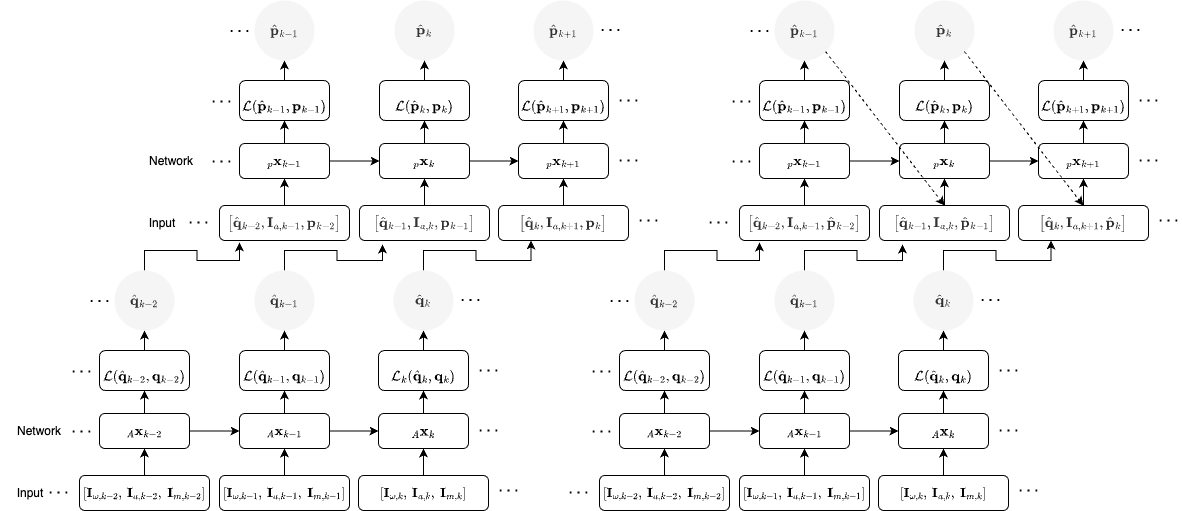}
\caption{ARIOT model training during first cycle (left) and second cycle (right). The $_{A}\mathbf{x}$ denotes the attitude neural network that are described in Section~\ref{sec:ariot} and $_{p}\mathbf{x}$ \ref{sec:riot}. The network inputs are given in Section~\ref{sec:sensors}, specifically Eq.~\ref{gyr}, \ref{acc} and \ref{mag}. Inference is performed in the similar depiction as the second cycle, without minimisation of the loss, detailed in Section~\ref{sec:infer}.}
\label{ariot}
\end{figure*}
We follow by parameterising the attitude in quaternions. Quaternions, which are a type of representation of attitudes in $\mathbb{R}^4$, have several advantages over representations in $\mathbb{R}^3$. They are free of discontinuities and singularities, and are more computationally efficient and numerically stable. To be a valid representation of an attitude, a quaternion must be a unit quaternion. Unit quaternions have a one-to-one correspondence with rotation matrices, and they double cover the group $\SO(3)$, meaning that both $\mathbf{q}$ and $-\mathbf{q}$ represent the same attitude. However, by requiring that $\mathbf{q}_0 \geq 0$, we can ensure that there is a unique correspondence between quaternions and rotation matrices \cite{kok2017using}.

We propose to use the self-attention mechanism and raw 9D IMU data in gradient descent optimisation to analyse and retain information related to gyroscope error and bias over long sequences. This minimises the complexity by forgoing preintegration. Additionally, the solution is unconstrained by not forcing the network into predefined dynamic models. These features and the inclusion of magnetometer measurements also have the advantage of the network being an an out-of-the-box solution where the local magnetic field is known. 

\subsubsection{Loss Function}\label{sec:quatdescription} 
We propose a new loss function for quaternions, which we call the Quaternion Loss. To define the Quaternion Loss, we first introduce some quaternion background and notation. A quaternion is a 4-tuple $(x, y, z, w)$, where $x, y, z, w$ are real numbers. Quaternions can be represented in the form
\begin{equation}
\mathbf{q} = w + xi + yj + zk
\end{equation}
where $i, j, k$ are the imaginary units, satisfying $i^2 = j^2 = k^2 = ijk = -1$.

Quaternions can be used to represent rotations in three-dimensional space by setting $w$ to the cosine of the rotation angle and $x, y, z$ to the sine of the rotation angle, multiplied by the rotation axis \cite{huynh2009metrics}. Given a pair of quaternions $(\mathbf{q}_1, \mathbf{q}_2)$, we can measure the similarity between them using the inner product as

\begin{equation}
\langle \mathbf{q}_1, \mathbf{q}_2\rangle = x_1x_2 + y_1y_2 + z_1z_2 + w_1w_2 
\end{equation} 

This product is related to the angle between the quaternions by the following,
\begin{equation}
\cos(\theta) = \frac{\langle \mathbf{q}_1, \mathbf{q}_2\rangle }{|\mathbf{q}_1| \cdot |\mathbf{q}_2|}
\end{equation} 
where $\theta$ is the angle between the quaternions and $| \cdot |$ denotes the L2 norm.

We then define the Quaternion Loss function as
\begin{equation}
\mathcal{L}(\mathbf{q}_1, \mathbf{q}_2) = \cos^{-1}(\operatorname{clamp}(\langle \mathbf{q}_1, \mathbf{q}_2\rangle, -1 + \epsilon, 1 - \epsilon))
\end{equation} 
where 
$$\operatorname{clamp}(x, a, b) = \begin{cases} a & \text{if } x < a \\ x & \text{if } a \leq x \leq b \\ b & \text{if } x > b \end{cases}$$
and $\epsilon$ is a small positive constant used to avoid numerical instability when the inner product is outside of the range $[-1, 1]$.\\

The mean angle across the batch is then returned as
\begin{equation}
\mathcal{L} = \frac{1}{N} \sum_{i=1}^N \theta_i 
\end{equation} 
where $N$ is the batch size.\\

\subsection{Recursive Inertial Odometry Transformer}
\label{sec:riot}
\begin{figure*}
\centering
\includegraphics[width=0.99\textwidth]{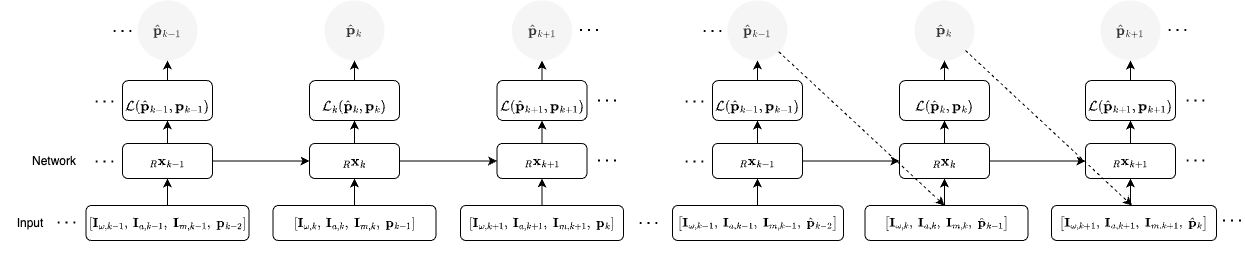}
\caption{RIOT model training during first cycle (left) and second cycle (right). The $_{R}\mathbf{x}$ denotes the attitude neural network described in Section~\ref{sec:riot}. The network inputs are given in Section~\ref{sec:sensors}, specifically Eq.~\ref{gyr}, \ref{acc} and \ref{mag}. Inference is performed in the similar depiction as the second cycle, without minimisation of the loss, detailed in Section~\ref{sec:infer}.}
\label{riot}
\end{figure*}

The Recursive Inertial Odometry Transformer is a self-attention based encoder-decoder network. Forgoing the attitude module to directly apply self-attention to raw 9D IMU data (from Eq.~\ref{gyr}, Eq.~\ref{acc} and Eq.~\ref{mag}) in gradient descent optimisation for 3D displacement regression; depicted in Figure.~\ref{riot}. 

The input to the network is a concatenation of the inertial measurements and true position priors in the first cycle of training; then true position priors are replaced by estimated position priors. The input is passed through an embedding layer to generate embedded representations. The encoder then applies self-attention to compute a weighted sum of the embedded representations for each time step, which is used to compute a context vector. The context vector is then passed through a decoder to estimate the 3D position at each time step. The equations for the input sequence, the embedding function, and the self-attention mechanism are provided in Section~\ref{network}. The model is then trained to minimise the Mean Square Error (MSE) loss function in Eq.~\ref{losspos} using the ADAM optimisation algorthim \cite{kingma2014adam}.

\begin{equation}\label{losspos}
\mathcal{L}(\hat{\mathbf{p}}, \mathbf{p})=\frac{1}{N T} \sum_{n=1}^N \sum_{t=1}^T\left\|\hat{\mathbf{p}}_{n, t}-\mathbf{p}_{n, t}\right\|^2
\end{equation}
where $\left\| \cdot \right\|^2$ represents the squared Euclidean norm, $N$ is the batch size, $T$ the sequence length and $\hat{\mathbf{p}}$ and $\mathbf{p}$ are the estimated and true positions, respectively. 

\section{Evaluation}
Despite numerous proposed solutions in literature attempt to solve inertial navigation, these approaches evaluate their algorithms using their datasets with various preprocessing and alignment techniques, such as the Umeyama algorithm \cite{umeyama1991least}. Under these conditions, it is difficult to compare directly to these different algorithms. Additionally, to the best of our knowledge, no other approach leverages all available IMU information. However, the inclusion of magnetometer measurements come with the drawback of our solutions being dependent on the local magnetic field, as the magnetometer readings are used to disambiguate the orientation of the IMU. This results in our network calibrations being regional specific and not generalisable to other datasets without local magnetic field knowledge. To this end, we build on our own implementation of a RNN as a means of comparison.

\subsection{Gated Recurrent Unit}\label{sec:gru}
Recent work on RNNs have shown that a Gated Recurrent Unit (GRU) surpasses the preferred LSTM in a number of scenarios \cite{yang2020lstm, gruber2020gru, cahuantzi2021comparison}. Additionally, GRUs have fewer parameters making it more computationally efficient, and has been shown to be more robust to noise and missing data \cite{che2018recurrent}. 

\begin{figure*}[h]
\centering
\includegraphics[width=0.99\textwidth]{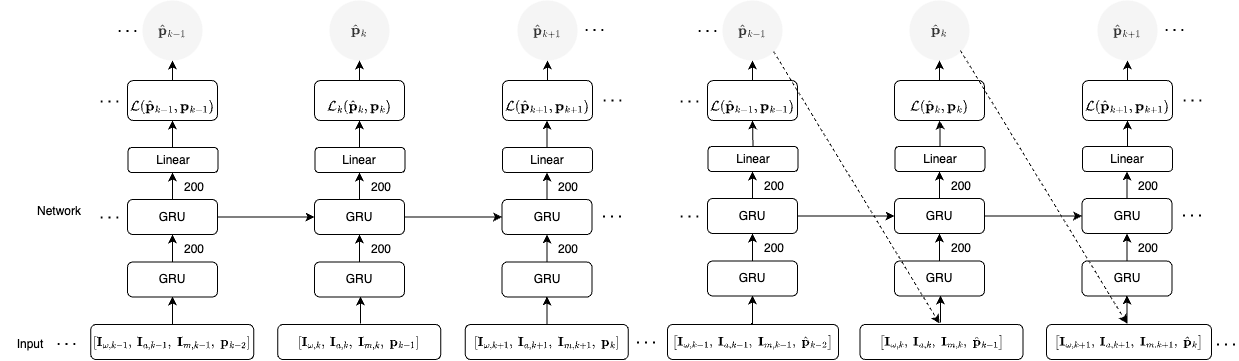}
\caption{GRU model training during first cycle (left) and second cycle (right). Out model is made up of 2 GRU cells, described in Section~\ref{sec:gru}, with 200 neurons per layer. The network inputs are given in Section~\ref{sec:sensors}, specifically Eq.~\ref{gyr}, \ref{acc} and \ref{mag}. Inference is performed in the similar depiction as the second cycle, without minimisation of the loss, detailed in Section~\ref{sec:infer}.}
\label{grufig}
\end{figure*}

We have added our own implementation of 2-layer GRU as a means of comparison. GRU has been shown to effective in inertial attitude estimation \cite{weber2021riann}, however to the best of our knowledge, GRUs are untested in inertial odometry domain. The network is formulated with the hidden state $h_t$ at time step $t$ as follows 

\begin{equation}\label{gru1}
r_t = \sigma(\mathbf{W}_{ir} x_t + \mathbf{b}_{ir} + \mathbf{W}_{hr} h_{t-1} + \mathbf{b}_{hr})
\end{equation}
\begin{equation}\label{gru2}
z_t = \sigma(\mathbf{W}_{iz} x_t + \mathbf{b}_{iz} + \mathbf{W}_{hz} h_{t-1} + \mathbf{b}_{hz})
\end{equation}
\begin{equation}\label{gru3}
\tilde{h_t} = \text{LeakyRe}LU(\mathbf{W}_{ix} x_t + \mathbf{b}_{ix} + r_t * (\mathbf{W}_{hx} h_{t-1} + \mathbf{b}_{hx}))
\end{equation}
\begin{equation}\label{gru4}
h_t = (1 - z_t) * h_{t-1} + z_t * \tilde{h_t}
\end{equation}
where $x_t$ is the input at time step $t$, $\mathbf{W}_{i*}$ and $\mathbf{b}_{i*}$ are the input-to-hidden weights and biases, $\mathbf{W}_{h*}$ and $b_{h*}$ are the hidden-to-hidden weights and biases, and $\sigma$ is the sigmoid function. Eq.~\ref{gru1} and Eq.~\ref{gru2} compute the reset gate, $r_t$, and update gate, $z_t$, respectively. These gates control the amount of information that is passed through to the next time step. Eq.~\ref{gru3} equation computes the candidate hidden state $\tilde{h_t}$, and Eq.~\ref{gru4} updates the hidden state $h_t$ by combining the previous hidden state $h_{t-1}$ and the candidate hidden state $\tilde{h_t}$.

We implement this network in the same manner as RIOT, depicted in Figure.~\ref{grufig}, where a stack of two GRU layers transforms the $9$D IMU input at sampling instant $t$, concatenated with the $3$D position vector at time $t-1$, to an $N_n$-dimensional feature vector $h_t$, with $N_n=200$ being the number of neurons per layer.

\subsection{Training and Dataset}
This approach was trained and tested on a publicly available smartphone data, published by Chen et al. \cite{OXIOD}. The dataset contains 158 sequences, totalling more than 42 km in total distance and incorporates a variety of attachments, activities and users to best reflect the broad use cases seen in real life. The data was captured via five different users and four different types of off-the-shelf consumer smartphones. The IMU data was collected and synchronised with a frequency of $100$ Hz, which is generally accepted in various applications and research~\cite{vleugels2021ultra, girbes2021asynchronous, dey2022function}. A high precision optical motion capture system (Vicon) was used to capture full pose ground truth at $0.01m$ location and $0.1$ degree attitude accuracy \cite{vicon}. The dataset was randomly divided into training, validation and test sets, following \cite{goodfellow2016deep}. A single sequence was left out for each of the variables as a means of complete, unseen comparison with other techniques. To avoid overfitting and to improve compute efficiency, we used a sliding window to capture $100$ measurements every $50$ to feed into the encoder and used random search to tune the hyperparameters. This gave us $63,614$ training samples, $18,175$ validation samples and $9089$ test samples. The implementation of all adaptations was carried out with PyTorch. The attitude network converges after $300$ epochs. The position network converges after $120$ and $30$ epochs, using true and recursive inputs, respectively. The learning rate of $0.001$, an ADAM optimiser and a dropout of $0.2$ was used across each implementation. The training was conducted in parallel on $4\times$ Nvidia V100 GPUs.

\subsection{Inference}
\label{sec:infer}
\begin{figure*}[h]
\centering
\includegraphics[width=0.99\textwidth]{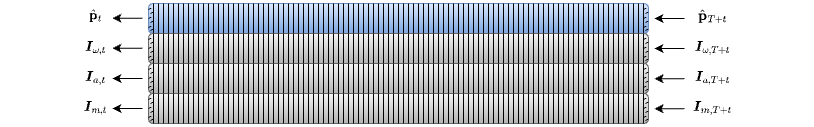}
\caption{Schematic of the sliding window recursive inference process for the RIOT model. The bottom three inputs are given in Section~\ref{sec:sensors}, specifically Eq.~\ref{gyr}, \ref{acc} and \ref{mag} and the top is the network position estimate, $\hat{\mathbf{p}}$ where $T$ is the sequence length and $t$ is the time step.}
\label{infer}
\end{figure*}

The inference procedure for each model closely resembles the second training cycle, as illustrated in Figures~\ref{ariot}, \ref{riot} and \ref{grufig} for ARIOT, RIOT, and the GRU, respectively. The initial window of each sequence is pre-padded with zeroes, followed by a given initial position. The position window is then iteratively updated by processing the subsequent windows of data. This inference approach is designed to reflect both recurrent architectures and recursive mathematical models, whilst leveraging the benefits of self-attention. This process is visualised in Figure~\ref{infer}.

\subsection{Evaluation Metrics}
\label{sec:loss}
In order to quantitatively assess the performance of each approach on each unobserved sequence of length $K$, the following three metrics were employed:\\

\begin{itemize}
\item \textbf{Absolute Trajectory Error (ATE)} $(\mathrm{m})$\\
\begin{equation} 
\text{ATE} = \sqrt{\frac{1}{K}\sum^K_{k=1} \left\|\left(\hat{\mathbf{p}}_k - \mathbf{p}_k\right)\right\|^2}
\end{equation}
The ATE is commonly used to assess the performance of a guidance or navigation system and represents the global accuracy of the estimated position. \\

\item \textbf{Relative Trajectory Error (RTE)}
$(\mathrm{m})$\\
\begin{equation} 
\sqrt{\frac{1}{K}\sum^K_{k=1}\left\|\left(\hat{\mathbf{p}}_{k+\Delta t} + \mathbf{p}_{k+\Delta t}\right)-\left(\hat{\mathbf{p}}_k + \mathbf{p}_k\right)\right\|^2}
\end{equation}
The RTE is a measure of the difference between the estimated and true position at a given time, relative to the distance between the two positions. It is often used to quantify the location position consistency over a pre-defined duration $\Delta t$; $\Delta t=1 s$ in this work.\\

\item \textbf{Cumulative Distribution Function (CDF)}\\ 
\begin{equation} 
\int_0^e f(x) d x, \quad \boldsymbol{e}=\left[\frac{1}{K} \sum_{k=1}^K \sqrt{\left\|\left(\hat{\mathbf{p}}_k - \mathbf{p}_k\right)\right\|^2}\right]
\end{equation}
The CDF is the distribution function $f(x)$, used to characterise the distribution of a variable. In this context it is used to describe the probability that the error in the estimated position will be less than or equal to a certain value. $f(x)$ is the probability density function of the localisation error $\boldsymbol{e}$.
\end{itemize}
ATE and RTE are used in deep inertial odometry papers \cite{yan2019ronin,wang2022a2dio} and CDF is a common metric in indoor localisation research \cite{7835628}.

\subsection{Evaluation}
\label{sec:loss}
\begin{table*}[t]
\centering
\caption{2D Position error metric comparison. A complete sequence was left out of the training data for each variable in the dataset. This was done as a means of unseen comparison over full sequences, allowing for different user, activity and device evaluations as well an overlook at the generalisability of each approach. Note that each network is capable of producing a 3D position estimate, however as the data was largely taken on a level plane where the discrepancy in the $z$-axis is far smaller than the $x$-$y$ plane, the addition of the vertical dimension would skew the error metrics. The best performing model over each sequence and for each metric has been made bold. RIOT performs best under most conditions, however ARIOT tracks better during highly dynamic motion. \label{tab1}}
\begin{adjustbox}{width=0.99\textwidth}
\begin{tabular}{@{\extracolsep{4pt}}cccccccccccccccc@{}}
\toprule
\multicolumn{1}{c}{} & \multicolumn{2}{c}{\textbf{User 2}} & \multicolumn{2}{c}{\textbf{User 3}} & \multicolumn{2}{c}{\textbf{User 4}}& \multicolumn{2}{c}{\textbf{User 5}} & \multicolumn{2}{c}{\textbf{Pocket}} & \multicolumn{2}{c}{\textbf{Running}}\\
\cmidrule{2-13} 
\multicolumn{1}{c}{\textbf{Model}} & \textbf{ATE (m)} & \textbf{RTE (m)}   & \textbf{ATE (m)} & \textbf{RTE (m)}  & \textbf{ATE (m)} & \textbf{RTE (m)} & \textbf{ATE (m)} & \textbf{RTE (m)}   & \textbf{ATE (m)} & \textbf{RTE (m)}  & \textbf{ATE (m)} & \textbf{RTE (m)}  \\ 
\midrule
GRU			& 0.0796		& 0.0110	& 0.0692	& 0.0100	& 0.0757  & 0.0121  	& 0.0856		& 0.0114		& 0.1013	& 0.125		& 0.1589		& 0.0171		\\
ARIOT			& 0.0994			& 0.0093		& 0.0934		& 0.0088			& 0.0960		& 0.0094	& 0.1027		& 0.0100		& 0.1059		& 0.0088			& 0.1279		& 0.0144		\\
RIOT			& \textbf{0.0681}			& \textbf{0.0090}		& \textbf{0.0655}	& \textbf{0.0085}		& \textbf{0.0654}	& \textbf{0.0091} 	& \textbf{0.0721}	& \textbf{0.0096}		& \textbf{0.0676}	& \textbf{0.0085}			& \textbf{0.0990}		& \textbf{0.0140}		\\
\midrule
\multicolumn{1}{c}{} & \multicolumn{2}{c}{\textbf{Slow Walking}} & \multicolumn{2}{c}{\textbf{Trolley}} & \multicolumn{2}{c}{\textbf{Handbag}}& \multicolumn{2}{c}{\textbf{Handheld}} & \multicolumn{2}{c}{\textbf{iPhone 5}} & \multicolumn{2}{c}{\textbf{iPhone 6}}\\
\cmidrule{2-13} 
\multicolumn{1}{c}{\textbf{Model}} & \textbf{ATE (m)} & \textbf{RTE (m)}   & \textbf{ATE (m)} & \textbf{RTE (m)}  & \textbf{ATE (m)} & \textbf{RTE (m)} & \textbf{ATE (m)} & \textbf{RTE (m)}   & \textbf{ATE (m)} & \textbf{RTE (m)}  & \textbf{ATE (m)} & \textbf{RTE (m)}  \\ 
\midrule
GRU			& 0.2634		& 0.0077	& 0.0881	& 0.0116	& 0.2021 & 0.0112 	& 7.352	& 0.0357	& 0.1172	& 0.0138	& 0.1133		& 0.0110	\\
ARIOT				& 0.1082			&  0.0060		& 0.1033		& 0.0099	& 0.1096 		& 0.0091		& \textbf{0.3196} & 0.0129 & 0.1046	 & 0.0090 & 0.1036  & 0.0089 \\
RIOT			& \textbf{0.0660}			& \textbf{0.0058}		& \textbf{0.0690}	& \textbf{0.0096} 	& \textbf{0.0694}	& \textbf{0.0089} 	& 0.4438	& \textbf{0.0109}	& \textbf{0.0690}	& \textbf{0.0086}		& \textbf{0.0667}		& \textbf{0.0085}		\\ 
\bottomrule
\end{tabular}
\end{adjustbox}
\end{table*}

This works presents three approaches for evaluation, on unseen sequences from different users, devices and activities. The ATE and RTE evaluation results are qualified in Table.~\ref{tab1}, with the best performing approach for each sequence and metric highlighted in bold. In addition, a qualitative analysis was conducted on the model's output, which revealed a close correspondence between the predicted trajectories and the ground truth trajectories. This is depicted in Appendix~\ref{appendix:A}, which provides visualisations of the position estimates for each approach during the first and last minute of data. RIOT performed best overall with the lowest ATE and RTE values, with the exception when the IMU was handheld. When the IMU is handheld and has implied consistent dynamic motion, the attitude estimation module in ARIOT is beneficial as it can help to disambiguate the accelerometer measurements which are affected by both linear acceleration and gravity. This lead to a more accurate trajectory estimate. However, when the IMU is mostly stable or cyclic through the motion, the additional complexity of the attitude estimation module is redundant and actually hinders the performance. We hypothesise that the model overly leans on the attitude representation, which in only beneficial in highly dynamic scenarios.
\begin{figure}[h]
  \centering
  \includegraphics[width=0.55\linewidth]{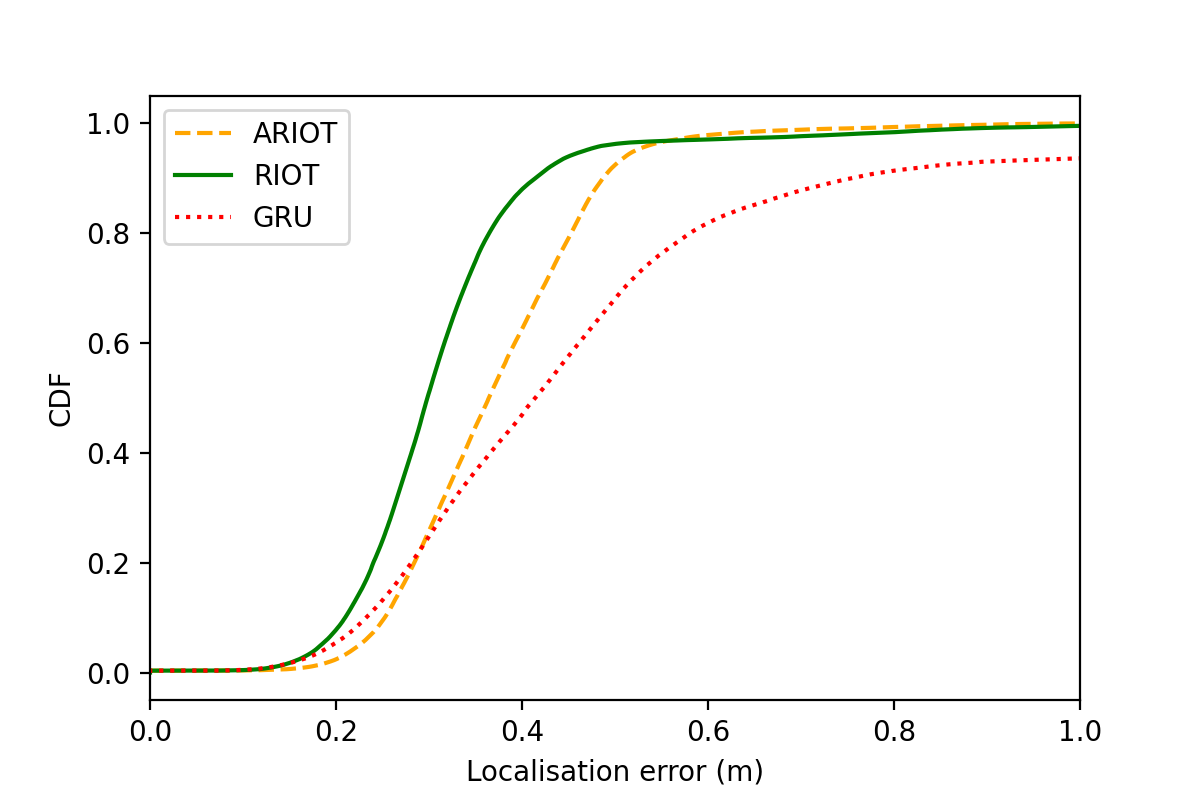}
  \caption{CDF of localisation error for each approach totalled over all test sequences. The CDF plots the percentage of localisation error values that fall below a certain percentage threshold, allowing evaluation for each approach at different levels of localisation error (m). RIOT performs best with a higher concentration of low errors in comparison to ARIOT and GRU.}
  \label{fig:cdf_plot}
\end{figure}

When analysing the performance of our models, it is important to consider the characteristics of the data and the specific scenario. We theorise the reason for the superior performance of RIOT is due to the simpler architecture, which is seemingly better suited for scenarios where the IMU is less dynamic. On the other hand, the additional complexity of ARIOT's attitude estimation module allows for improved handling of dynamic motion.

It is evident that the GRU performed considerably worse than both RIOT and ARIOT in all of the sequences. This is likely due to the fundamentally inferior design of the RNN model, leading to its inability to effectively process the complex motion present over long sequences. However, it is important to note that the GRU model still performed relatively well, which highlights the effectiveness of our learning process used in the development of the models. 

This analysis is further evidenced in Figure.~\ref{fig:cdf_plot}, which depicts the mean CDF of the localisation error over the total set of test sequences. RIOT performs almost consistently, indicated by the steep gradient of the CDF in the lower error range, whereas for the ARIOT and GRU, the errors are more spread out over a wider range of values.

\begin{figure}[!ht]
\centering
\begin{subfigure}{0.26\textwidth}
\includegraphics[width=\linewidth]{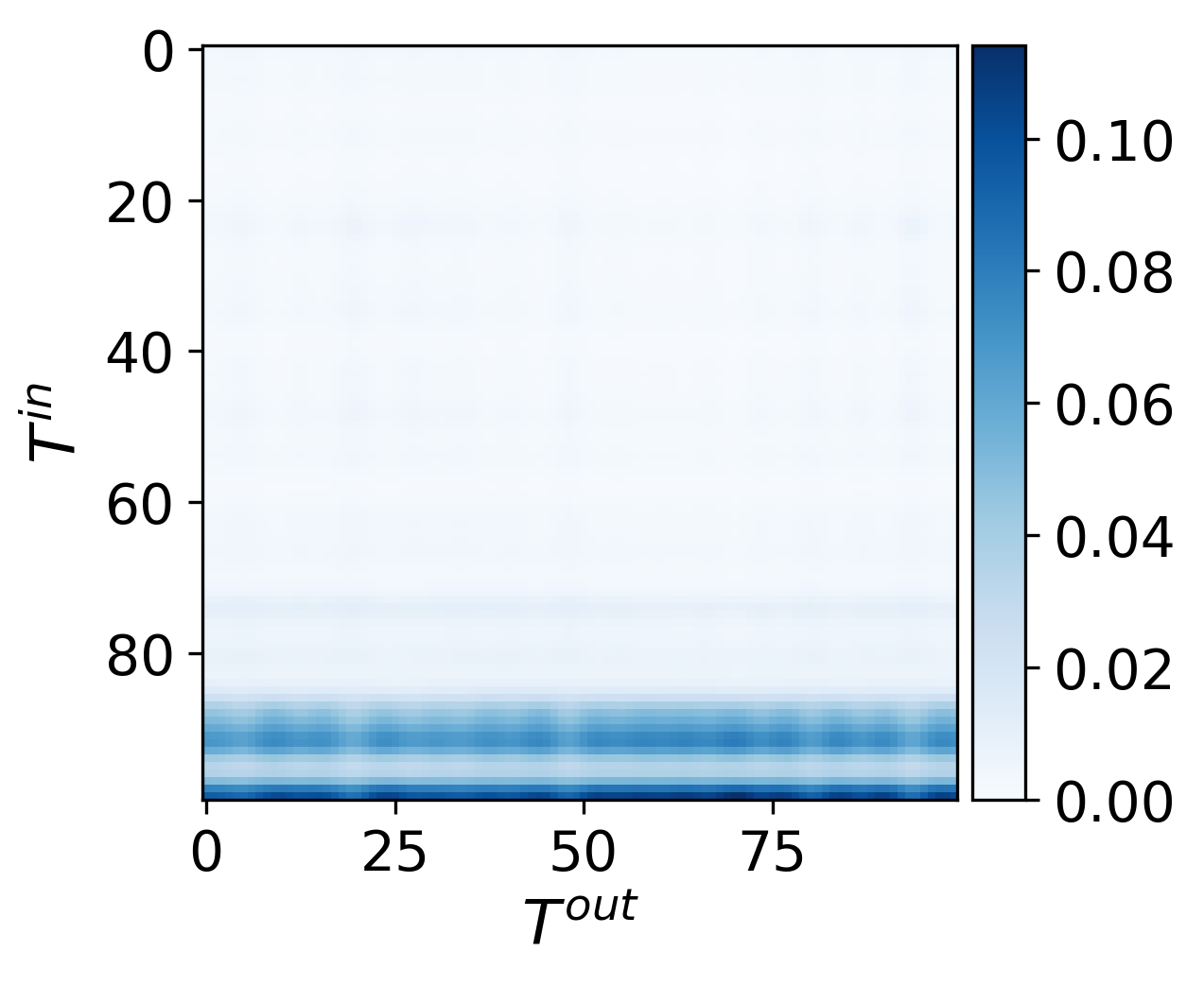}
\caption{}
\label{attn1a}
\end{subfigure}
\hspace{0.3cm}
\begin{subfigure}{0.465\textwidth}
\includegraphics[width=\linewidth]{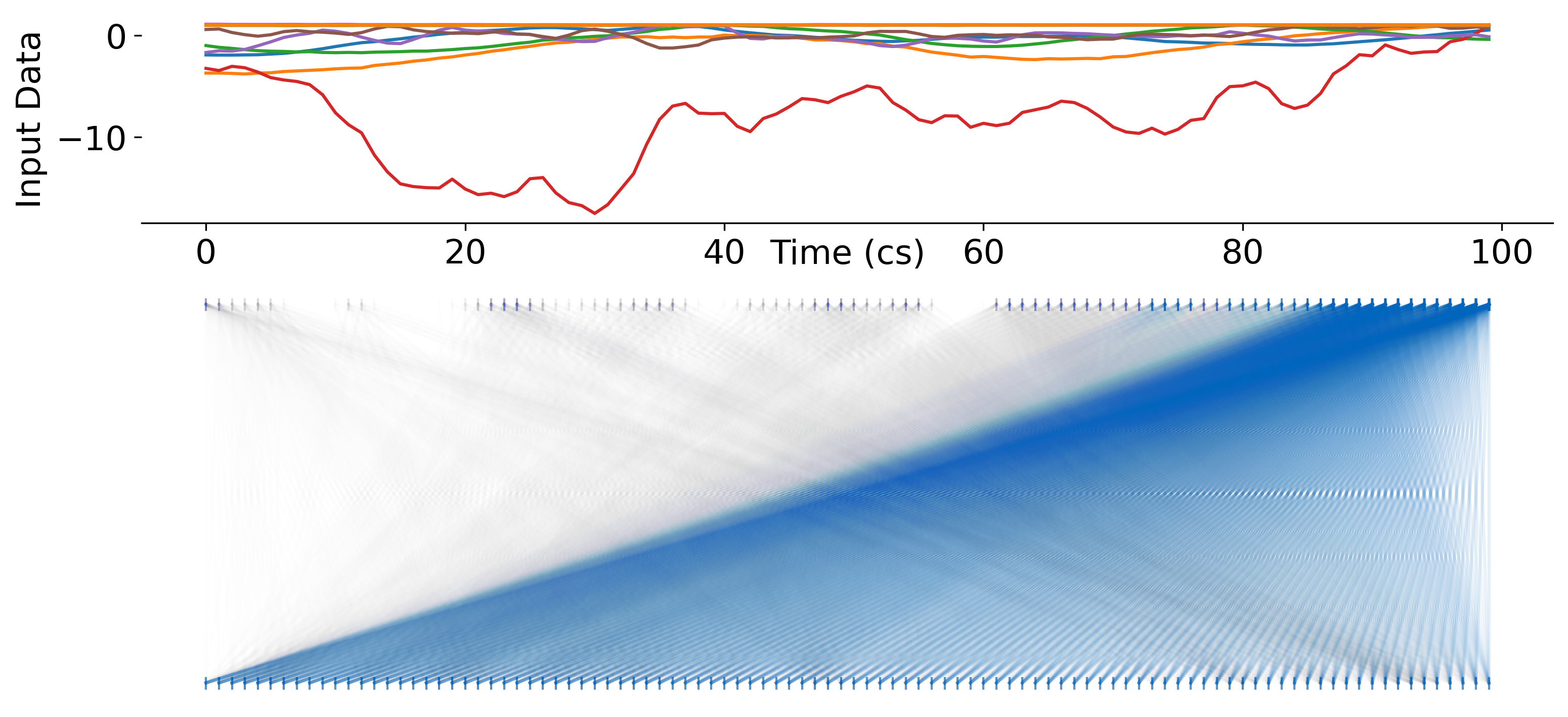}
\caption{}
\label{attn1b}
\end{subfigure}

\vspace{0.5cm}

\begin{subfigure}{0.26\textwidth}
\includegraphics[width=\linewidth]{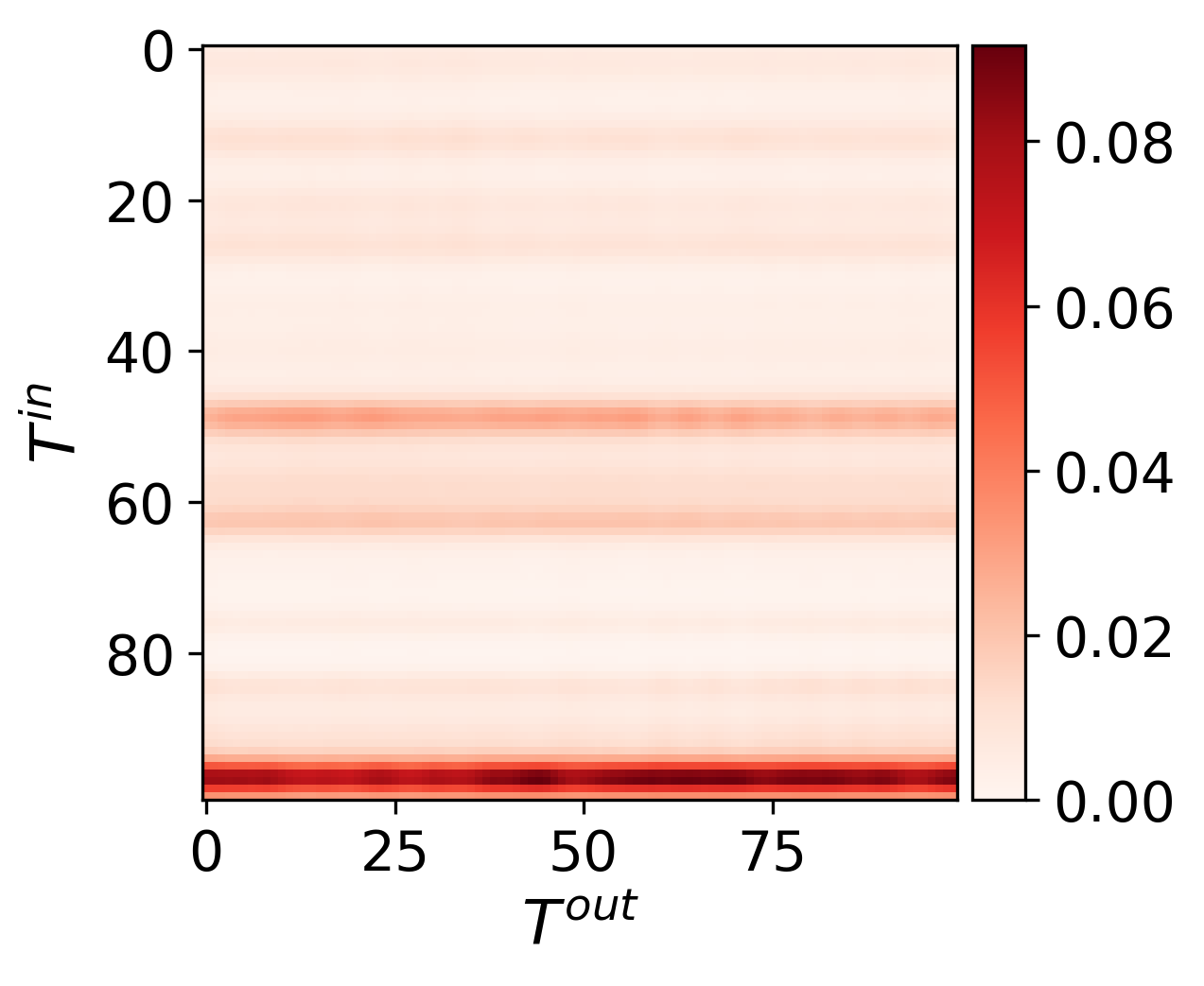}
\caption{}
\label{attn1c}
\end{subfigure}
\hspace{0.33cm}
\begin{subfigure}{0.465 \textwidth}
\includegraphics[width=\linewidth]{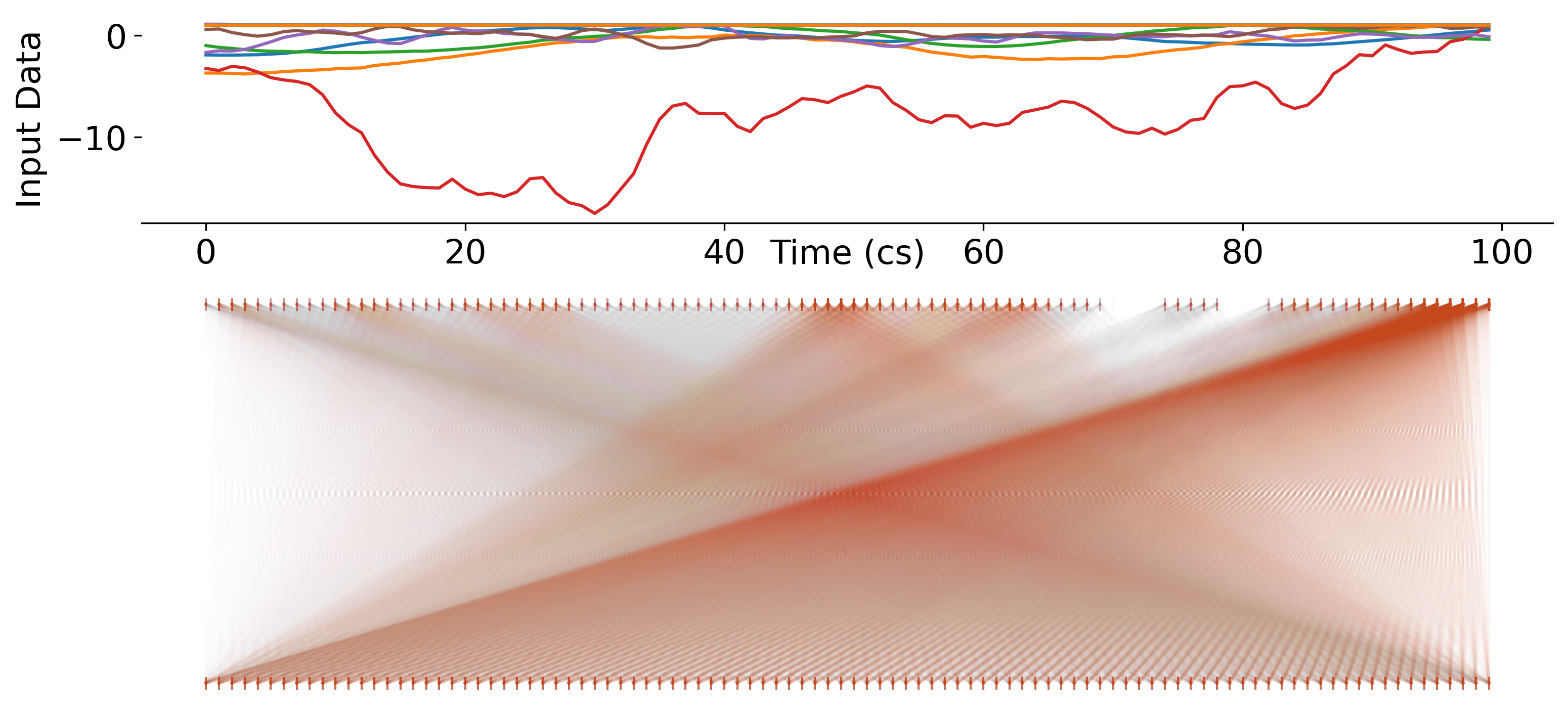}
\caption{}
\label{attn1d}
\end{subfigure}
\caption{Visualisations of the self-attention scores from the first encoder in RIOT on an arbitrary sequence of input data. \textbf{(a)} and \textbf{(b)} depict the attention scores from the first head of the first encoder as a matrix and bipartite graph, respectively. \textbf{(c)} and \textbf{(d)} depict the attention scores from the second head of the second encoder as a matrix and bipartite graph, respectively. \textbf{Left:} The heat matrix displays the attention scores assigned to each input element in a sequence. The darker the colour, the higher the attention weight given to that element, indicating that it has a greater impact on the final output. \textbf{Right:} The graph represents each input element as a node on one side of the graph, while the attention scores assigned by the model are represented as nodes on the other side. Edges connecting the nodes represent the attention weights, or the degree to which the model is considering each input element. The thickness of the edges represent the magnitude of the attention weights, with thicker edges indicating higher attention scores.}
\label{attn1}
\end{figure}

Our models utilise multi-headed self-attention, which is achieved through multiple parallel attention mechanisms. By allowing the model to attend to different parts of the input sequence dynamically, self-attention can capture complex relationships and dependencies. Each self-attention mechanism calculates an attention matrix $\boldsymbol{A}$ of size $T \times T$, where $T$ is the sequence length, by utilising the softmax operation as described in Eq.~\ref{eq:softmax}. The attention scores determine the influence of the input time features on the higher-level output time features.

\begin{figure}[!h]
\centering
\begin{subfigure}{0.26\textwidth}
\includegraphics[width=\linewidth]{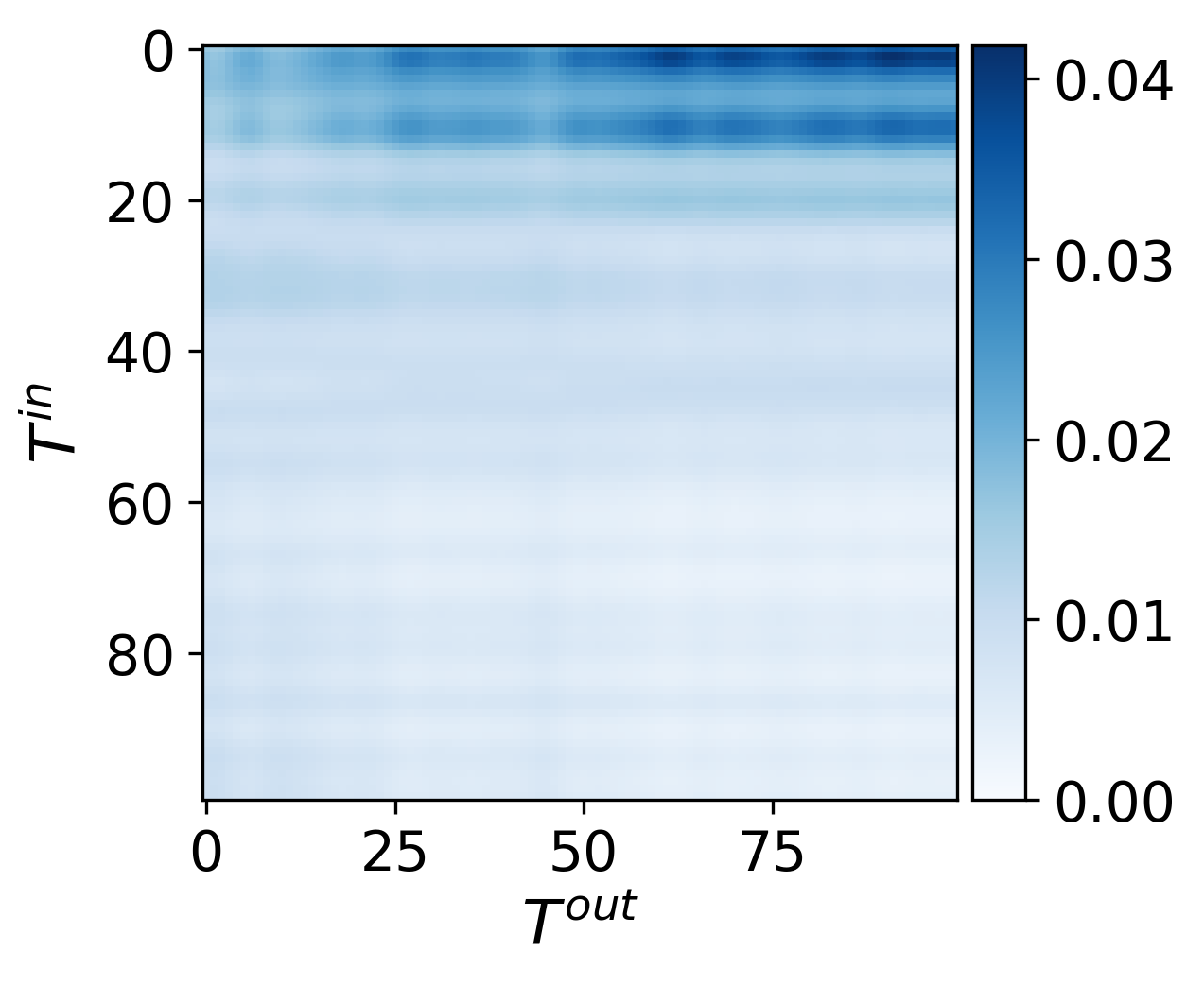}
\caption{}
\label{attn2a}
\end{subfigure}
\hspace{0.3cm}
\begin{subfigure}{0.465\textwidth}
\includegraphics[width=\linewidth]{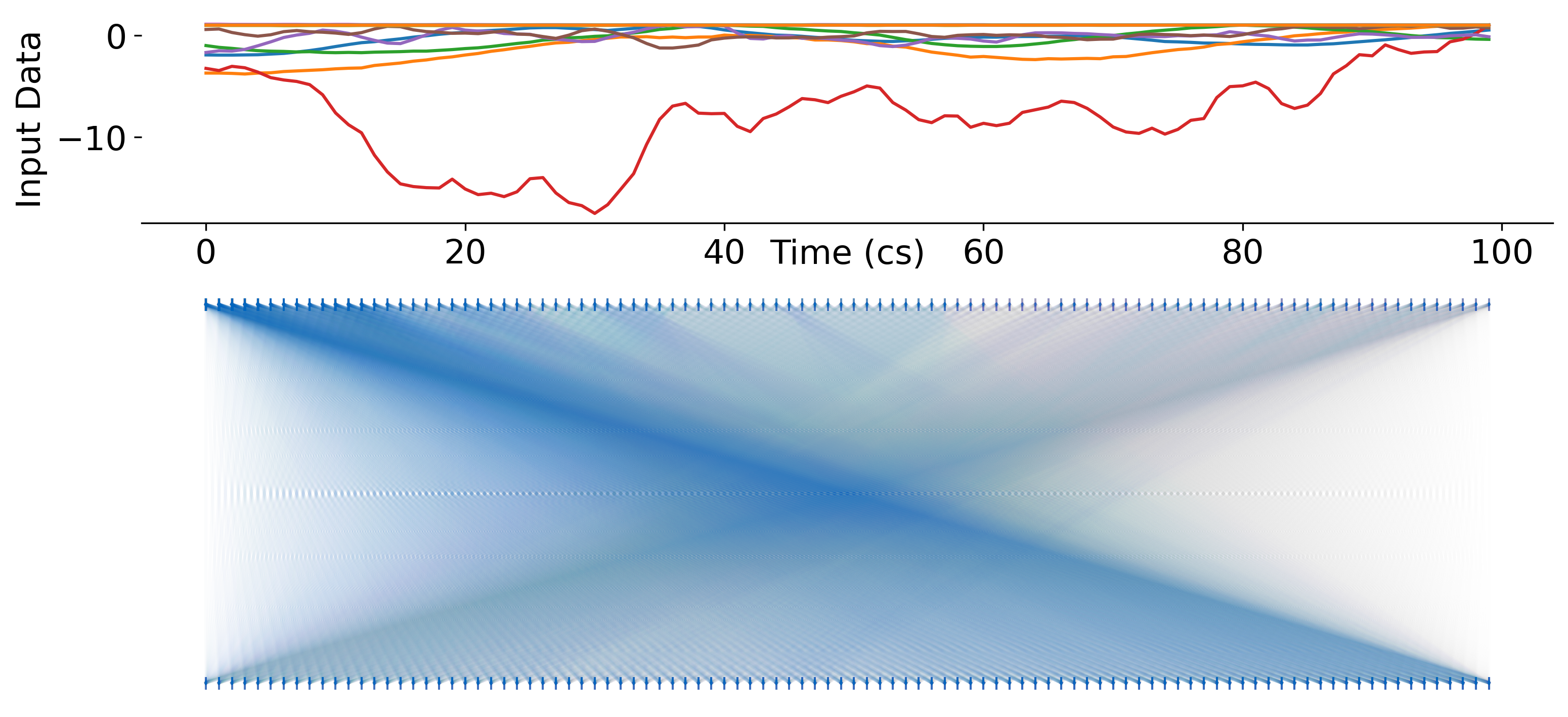}
\caption{}
\label{attn2b}
\end{subfigure}

\vspace{0.5cm}

\begin{subfigure}{0.26\textwidth}
\includegraphics[width=\linewidth]{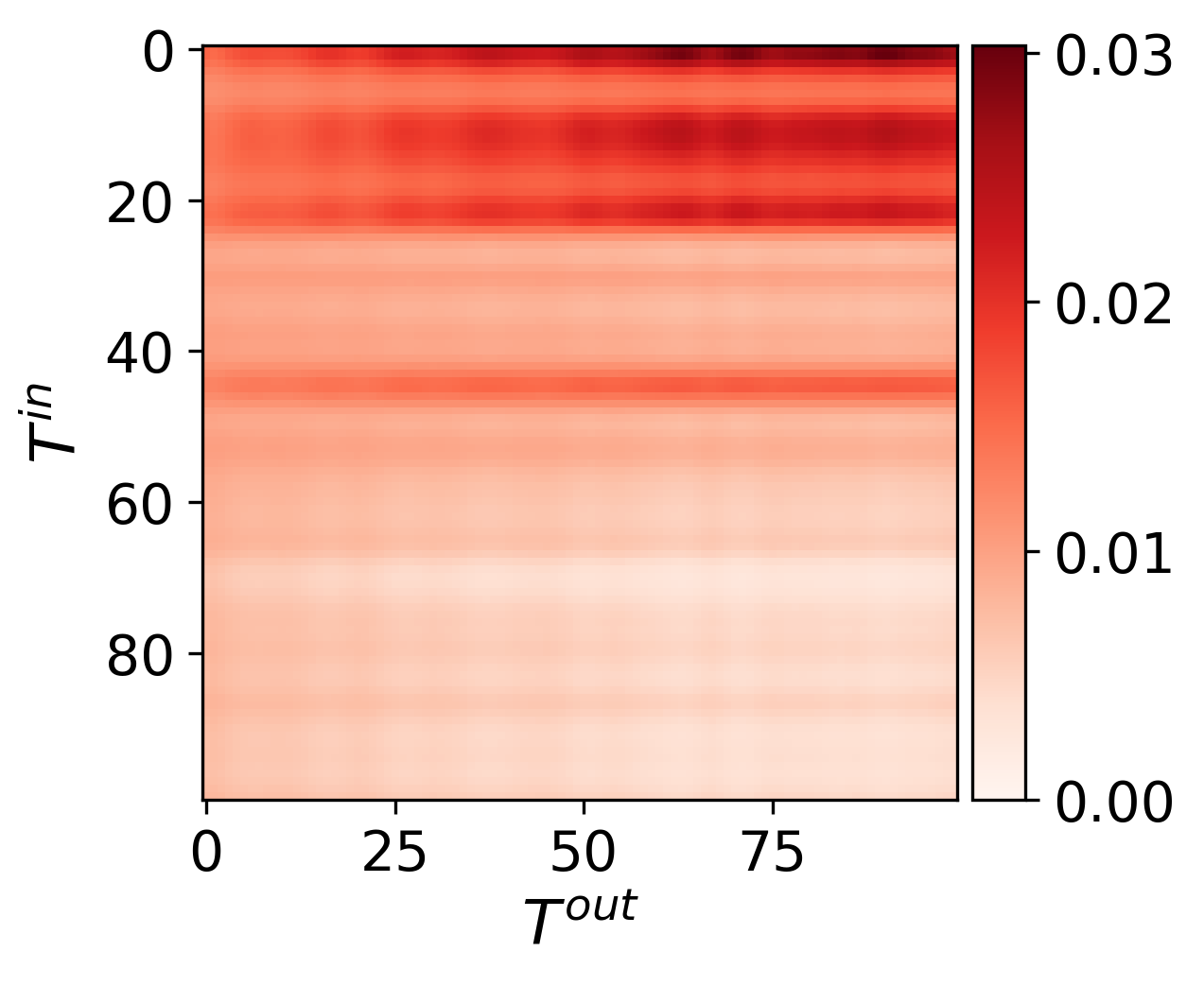}
\caption{}
\label{attn2c}
\end{subfigure}
\hspace{0.33cm}
\begin{subfigure}{0.465\textwidth}
\includegraphics[width=\linewidth]{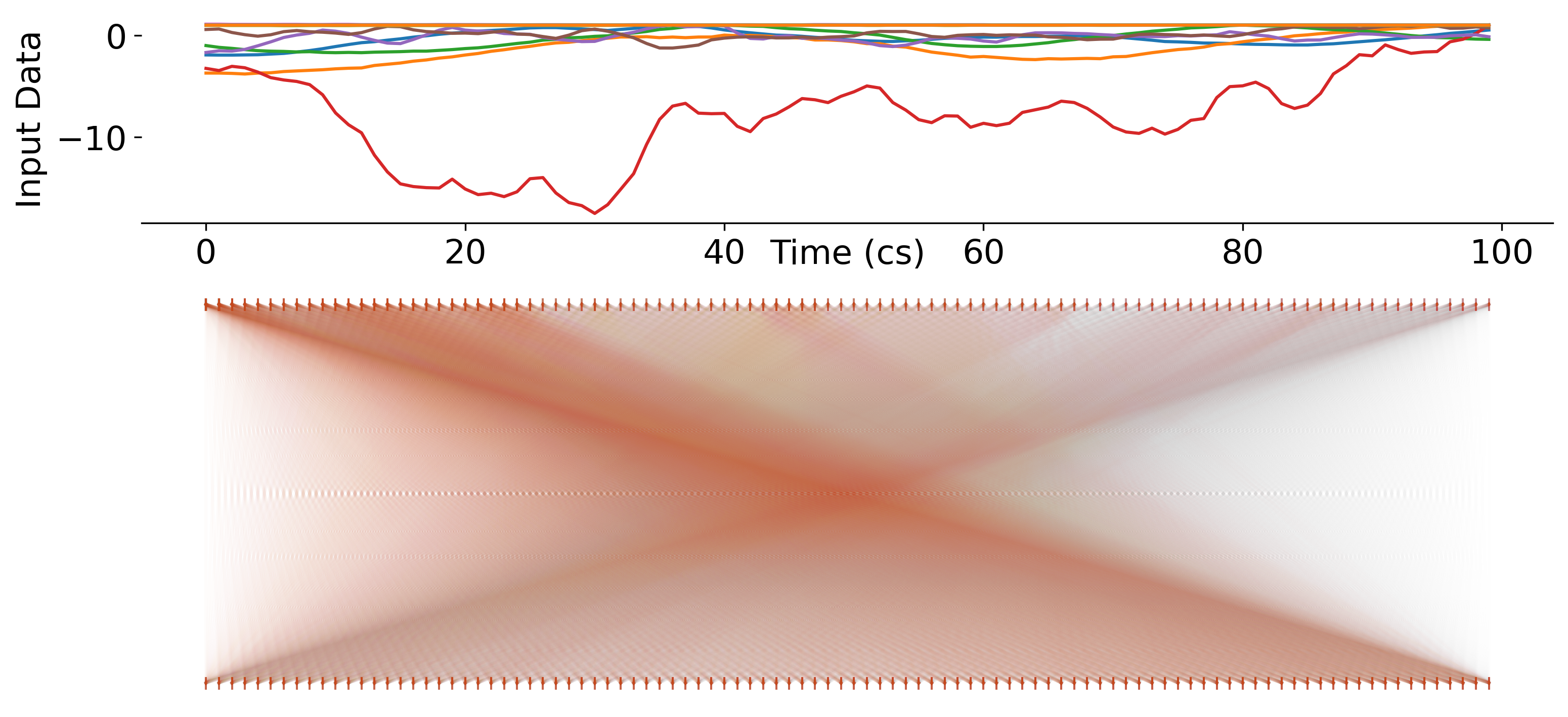}
\caption{}
\label{attn2d}
\end{subfigure}
\caption{Visualisations of the self-attention scores from the second encoder in RIOT on an arbitrary sequence of input data. \textbf{(a)} and \textbf{(b)} depict the attention scores from the first head of the second encoder as a matrix and bipartite graph, respectively. \textbf{(c)} and \textbf{(d)} depict the attention scores from the second head of the second encoder as a matrix and bipartite graph, respectively. \textbf{Left:} The heat matrix displays the attention scores assigned to each input element in a sequence. The darker the colour, the higher the attention weight given to that element, indicating that it has a greater impact on the final output. \textbf{Right:} The graph represents each input element as a node on one side of the graph, while the attention scores assigned by the model are represented as nodes on the other side. Edges connecting the nodes represent the attention weights, or the degree to which the model is considering each input element. The thickness of the edges represent the magnitude of the attention weights, with thicker edges indicating higher attention scores.}
\label{attn2}
\end{figure}

The matrix visualisations, in Figures~\ref{attn1a}, \ref{attn1c}, \ref{attn2a}, and \ref{attn2c}, provide a glimpse into how the model is weighing and combining multiple inputs to make a prediction. The values of the attention matrix depict two attention heads from the first self-attention layer from each encoder as an adjacency matrix between input nodes and output nodes. The matrix can also be represented as a bipartite graph, as shown in Figures~\ref{attn1b}, \ref{attn1d}, \ref{attn2b}, and \ref{attn2d}. The edge weights represent the strength of the attention, and the opacity of the edges indicates the magnitude. The input time series is shown above the attention graph as a reference, and the attention scores are depicted as vertical lines corresponding to the values in $\boldsymbol{A}$.

From the visual representations of the attention matrices, we can directly observe the distinctions between the different self-attention heads and encoders. The heads in the first encoder appear to be highly concentrated on the latter part of the sequence, whereas the heads in the second encoder concentrate on the beginning but have a greater overall attention. From Figures~\ref{attn1} and \ref{attn2} we observe that the model considers both short and longer-term temporal dependencies in the data when making predictions, rather than just focusing on the prior time step as seen in traditional methodologies. This is largely the reason for the accurate and stable position estimates, especially in situations where the motion is complex or noisy.\\

In summary, we evaluated the performance of three novel recursive deep inertial odometry frameworks. Our results show that self-attention based networks have superior performance over a GRU based RNN, with RIOT performing best overall, with a sequence length weighted mean ATE of $0.0865$m and RTE $0.0091$m. The mean RTE and ATE of ARIOT and GRU were $0.1134$m, $0.0095$m and $0.4594$m, $0.0130$m, respectively. Our results also revealed that a simpler architecture can generally yield better results, however having an attitude module dramatically improves performance in specific scenarios where the IMU experiences highly dynamic motion, highlighting the importance of evaluating solutions on diverse datasets.  \\

\section{RIOT Ablations}\label{ablations}
\textbf{Model Dimensionality.} \; \; \; We trained the model with differing dimensionality vector sizes from 56 to 896. Increasing the dimensionality of the model makes a small but measurable improvement up to until 224. This finding aligns with the general principle in deep learning in which complexity reaches a point to where passing it leads to overfitting, resulting in degraded performance on new data. 

\textbf{Encoder-Decoder Blocks.} \;\;\; Increasing the number of encoder-decoder blocks did not result in a decrease in the model's perplexity. We trained 3 different models, with 2, 4 and 6 blocks. 

\textbf{Attention Heads.} \; \; \; We trained the model with 2, 3 and 4 attention heads in each encoder-decoder block. There were small, almost immeasurable improvements in the networks performance on the test set. However,  when applied to the unseen sequences, the models with 3 or 4 heads performed considerably worse. We believe increasing the number of heads past 2 forces the network into overfitting.  

\textbf{Window Size.} \; \; \; We trained the model with differing window sizes from 50 to 500 (0.5 seconds to 5 seconds). As the window size incrementally increased over 100, we saw better test set results but worse results on the unseen sequences. Increasing the window size of the input data exponentially increases the model complexity, as RIOT has 12 input features. The added complexity forces the network into overfitting.

\section{Conclusion}
This work proposes novel self-attention based recursive neural network models, RIOT and ARIOT, for pose invariant inertial odometry. The proposed approaches incorporate true position priors in the training process and are trained on inertial measurements and ground truth displacement data, allowing for recursion and the ability to learn both motion characteristics and systemic error bias and drift. The evaluation results demonstrate that RIOT outperforms ARIOT and a GRU in terms of position error metrics, with a sequence length weighted mean Absolute Trajectory Error (ATE) of $0.0865$m and a sequence length weighted mean Relative Trajectory Error (RTE) of $0.0091$m. These results are significantly better than existing deep learning inertial odometry methods in literature, highlighting the effectiveness of the proposed approaches and learning methodology. Future work will consider the scalability of these approaches and making them local magnetic field agnostic.

\section*{Funding}
This research was undertaken with the assistance of resources from
the National Computational Infrastructure (NCI Australia), an
NCRIS enabled capability supported by the Australian
Government (Grant No. LE160100051).

\appendix
\section[\appendixname~\thesection]{Position estimate visulations from the first and last minute of each network}\label{appendix:A}
\begin{figure*}[t]
\centering
\begin{subfigure}{0.4\textwidth}
\includegraphics[width=\textwidth]{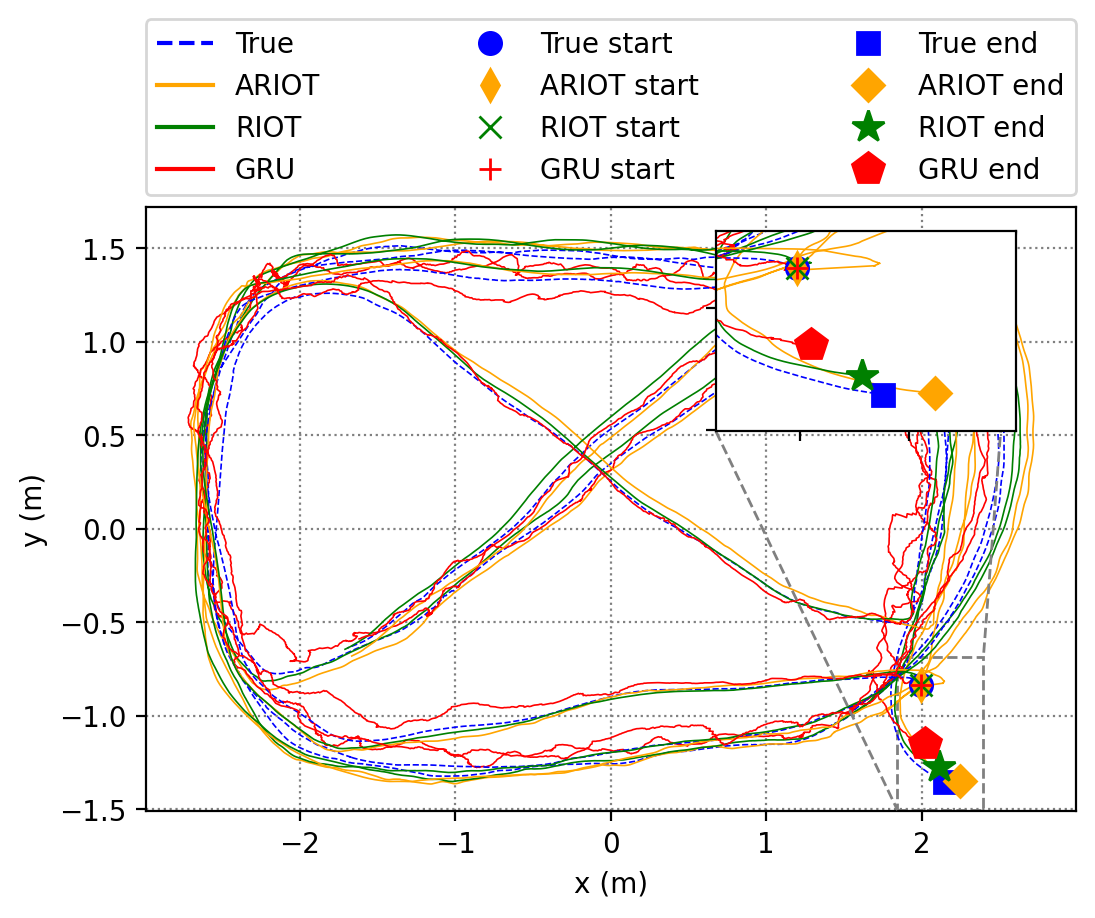}
\caption{User 2: Total Time: 263.5s, 
TRUE Distance: 220.9m, 
ARIOT Distance: 234.1m, 
RIOT Distance: 228.1m, 
GRU Distance: 260.7m}
\end{subfigure}
\hfill
\begin{subfigure}{0.4\textwidth}
\includegraphics[width=\textwidth]{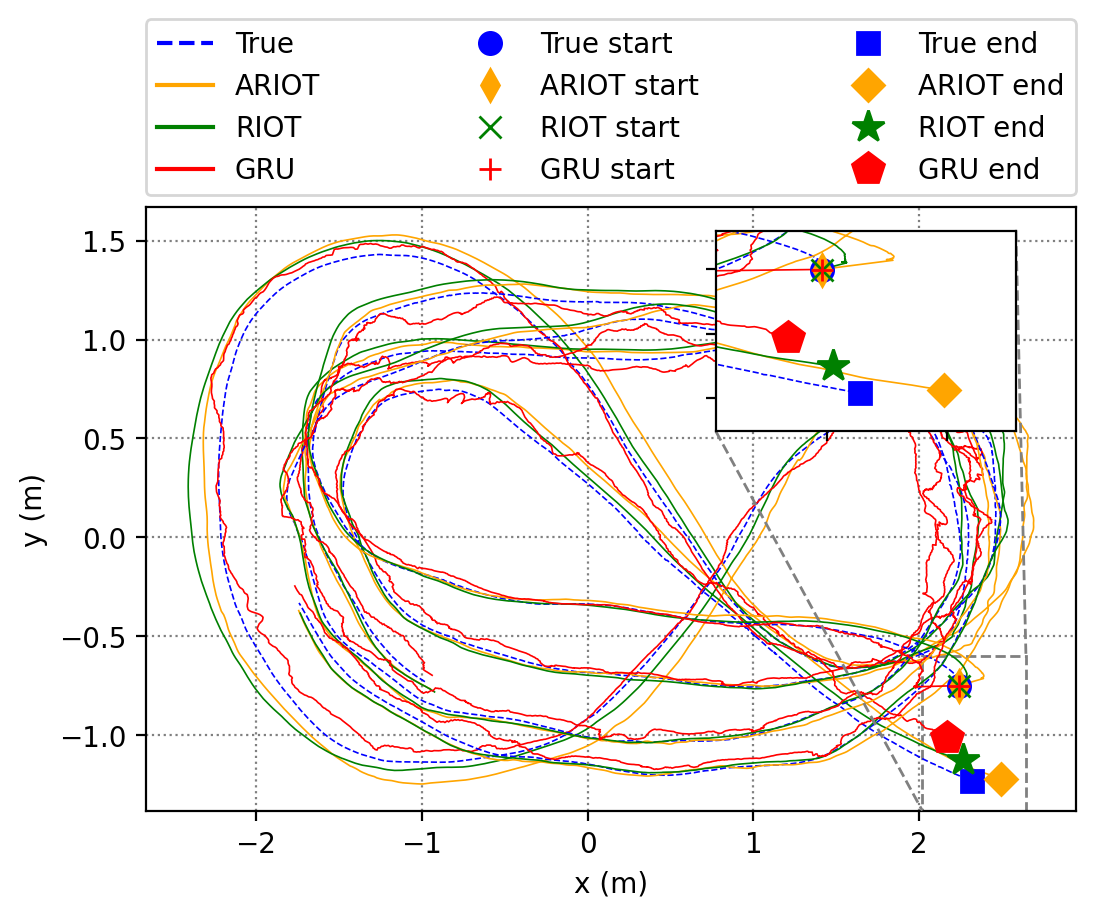}
\caption{User 3: Total Time: 258.0s, 
TRUE Distance: 203.3m, 
ARIOT Distance: 215.5m, 
RIOT Distance: 209.7m, 
GRU Distance: 229.3m}
\end{subfigure}

\vspace{1cm}

\begin{subfigure}{0.4\textwidth}
\includegraphics[width=\textwidth]{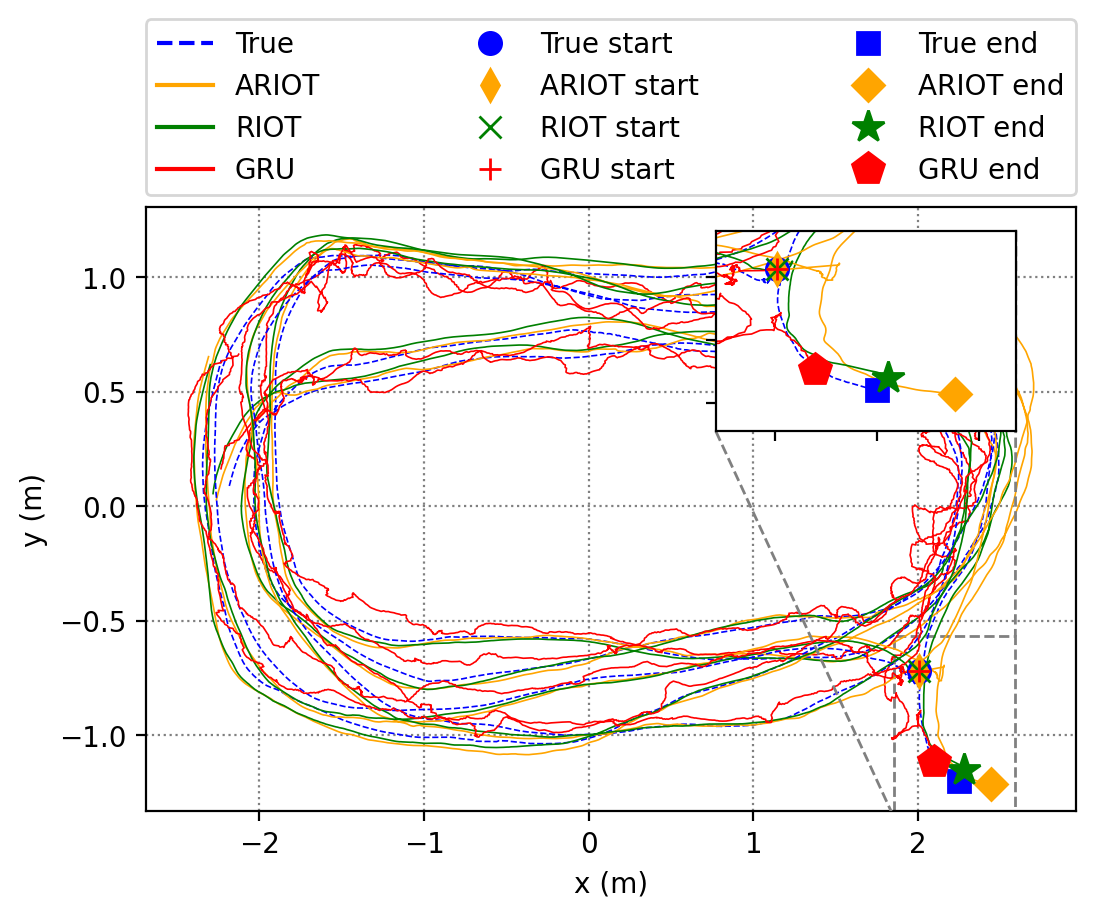}
\caption{User 4: Total Time: 434.7s, 
TRUE Distance: 364.3m, 
ARIOT Distance: 387.6m, 
RIOT Distance: 377.5m, 
GRU Distance: 448.6m}
\end{subfigure}
\hfill
\begin{subfigure}{0.4\textwidth}
\includegraphics[width=\textwidth]{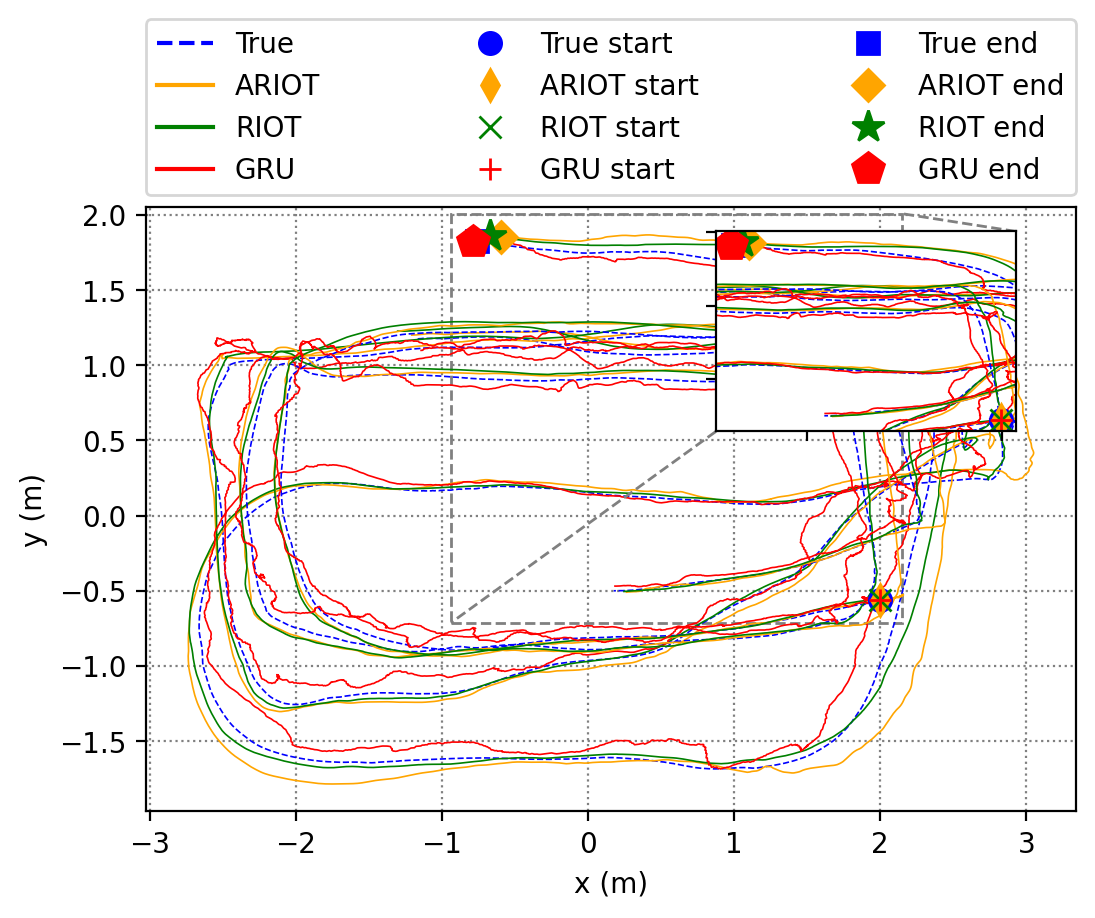}
\caption{User 5: Total Time: 152.0s, 
TRUE Distance: 130.3m, 
ARIOT Distance: 138.6m, 
RIOT Distance: 134.0m, 
GRU Distance: 153.7m}
\end{subfigure}

\vspace{1cm}

\begin{subfigure}{0.4\textwidth}
\begin{minipage}[t]{\textwidth}
\includegraphics[width=\textwidth]{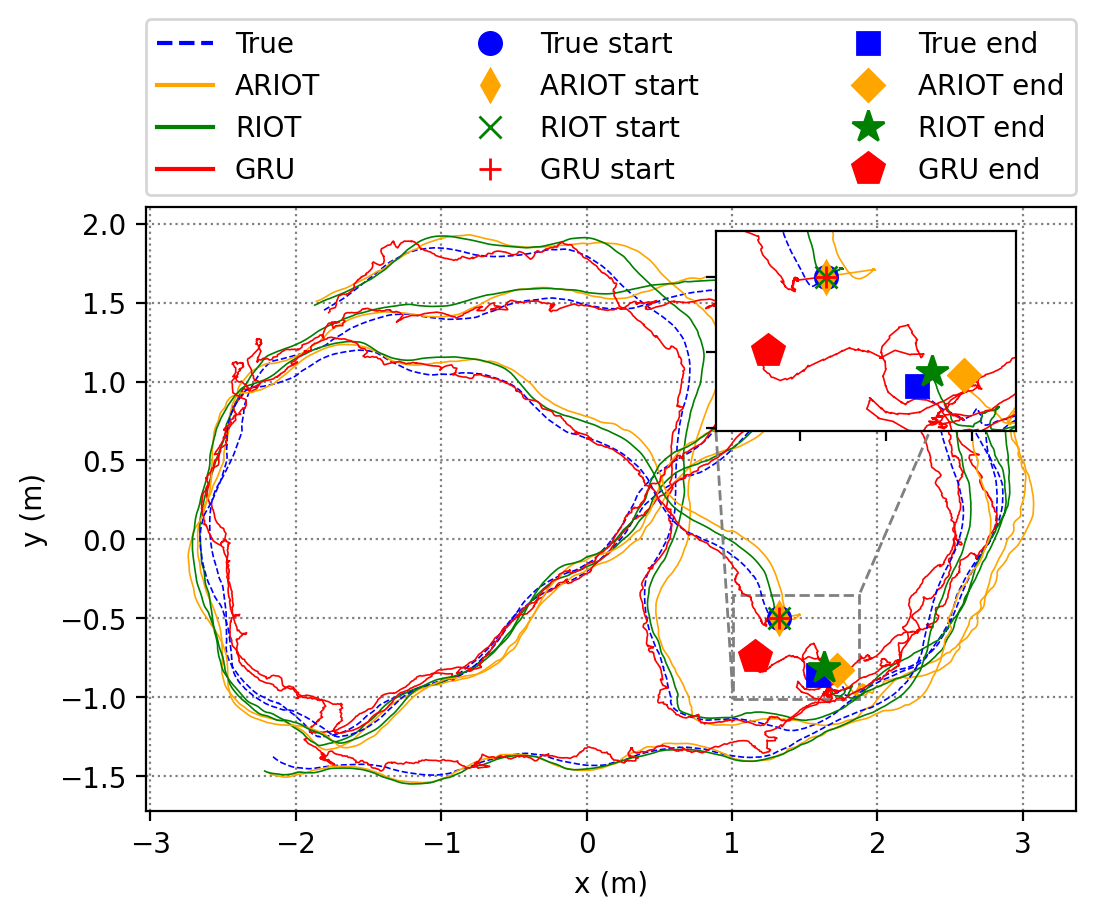}
\caption{Pocket: Total Time: 622.9s, 
TRUE Distance: 493.7m, 
ARIOT Distance: 526.3m, 
RIOT Distance: 512.5m, 
GRU Distance: 650.7m}
\end{minipage}
\end{subfigure}
\hfill
\begin{subfigure}{0.4\textwidth}
\begin{minipage}[t]{\textwidth}
\includegraphics[width=\textwidth]{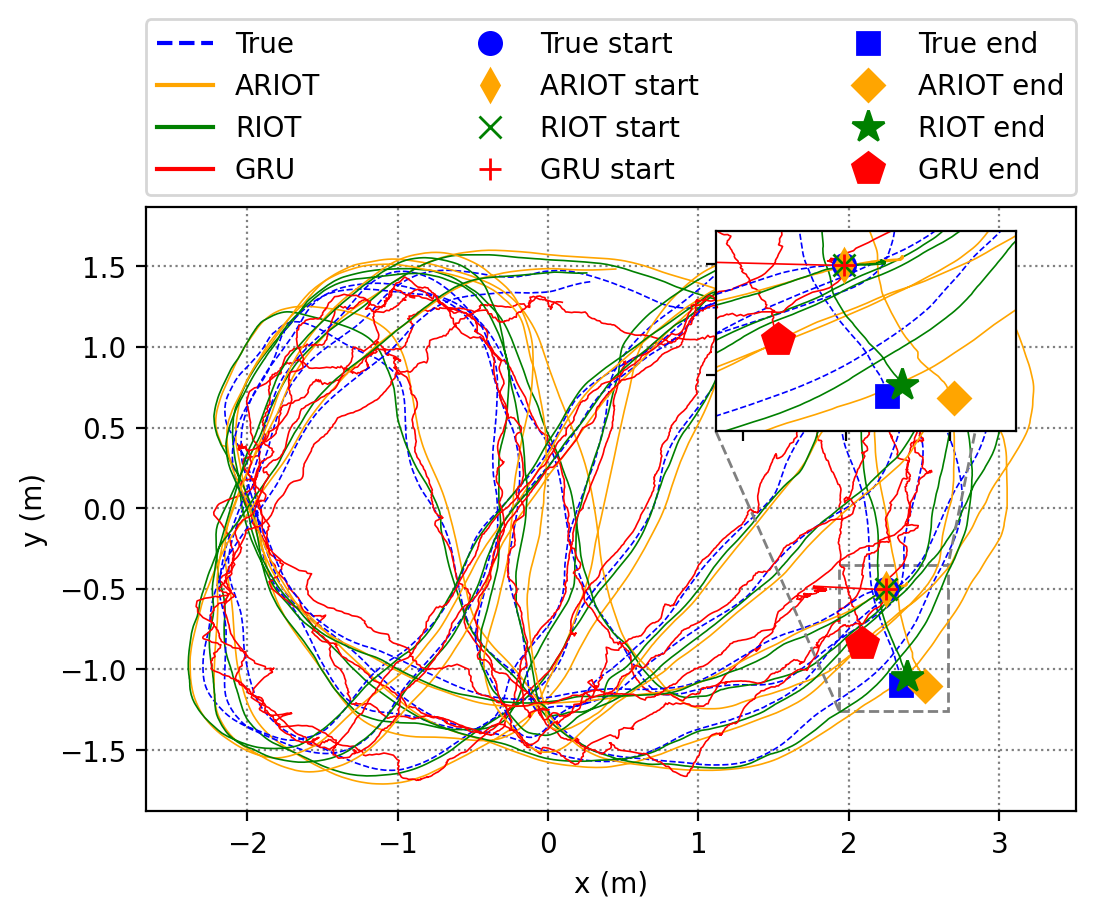}
\caption{Running: Total Time: 302.9s, 
TRUE Distance: 393.1m, 
ARIOT Distance: 422.2m, 
RIOT Distance: 410.5m, 
GRU Distance: 447.6m}
\end{minipage}
\end{subfigure}

\caption{Position estimates from the first and last minute of each network as well as the true path. The sequence run time is given as well as each approaches total path length.}
\label{fig:figures}
\end{figure*}

\begin{figure*}[t]
\centering
\begin{subfigure}{0.4\textwidth}
\begin{minipage}[t]{\textwidth}
\includegraphics[width=\textwidth]{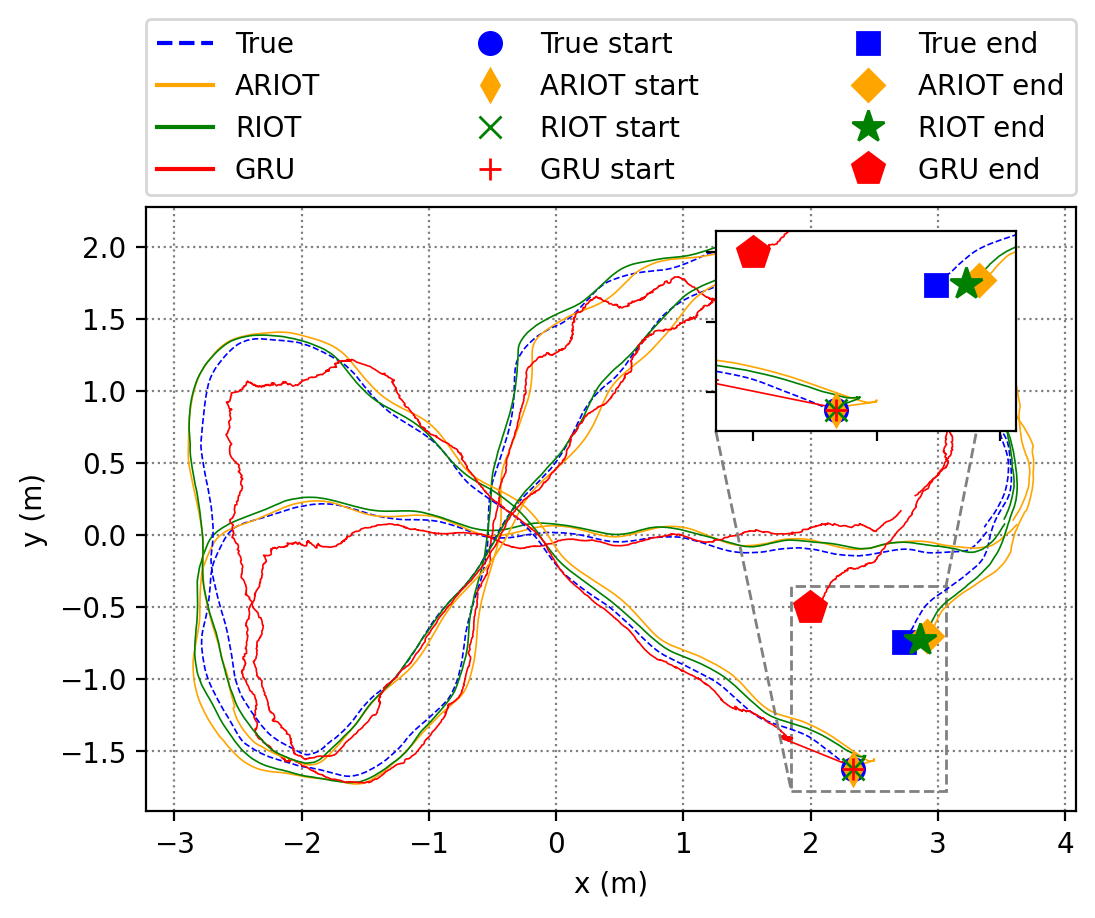}
\caption{Slow Walking: Total Time: 303.7s, 
TRUE Distance: 166.6m, 
ARIOT Distance: 174.7m, 
RIOT Distance: 171.1m, 
GRU Distance: 185.2m}
\end{minipage}
\end{subfigure}
\hfill
\begin{subfigure}{0.4\textwidth}
\begin{minipage}[t]{\textwidth}
\includegraphics[width=\textwidth]{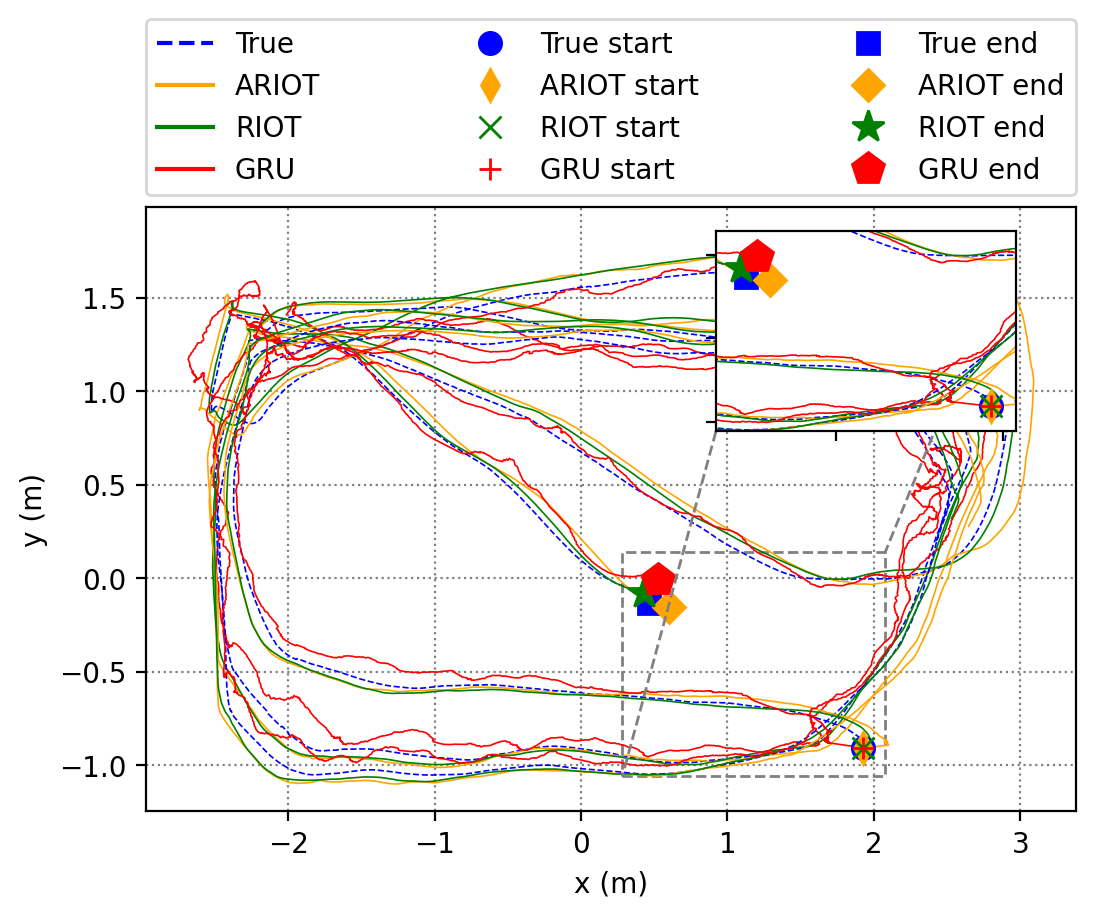}
\caption{Trolley: Total Time: 370.0s, 
TRUE Distance: 328.4m, 
ARIOT Distance: 346.5m, 
RIOT Distance: 337.4m, 
GRU Distance: 376.8m}
\end{minipage}
\end{subfigure}

\vspace{1cm}

\begin{subfigure}{0.4\textwidth}
\begin{minipage}[t]{\textwidth}
\includegraphics[width=\textwidth]{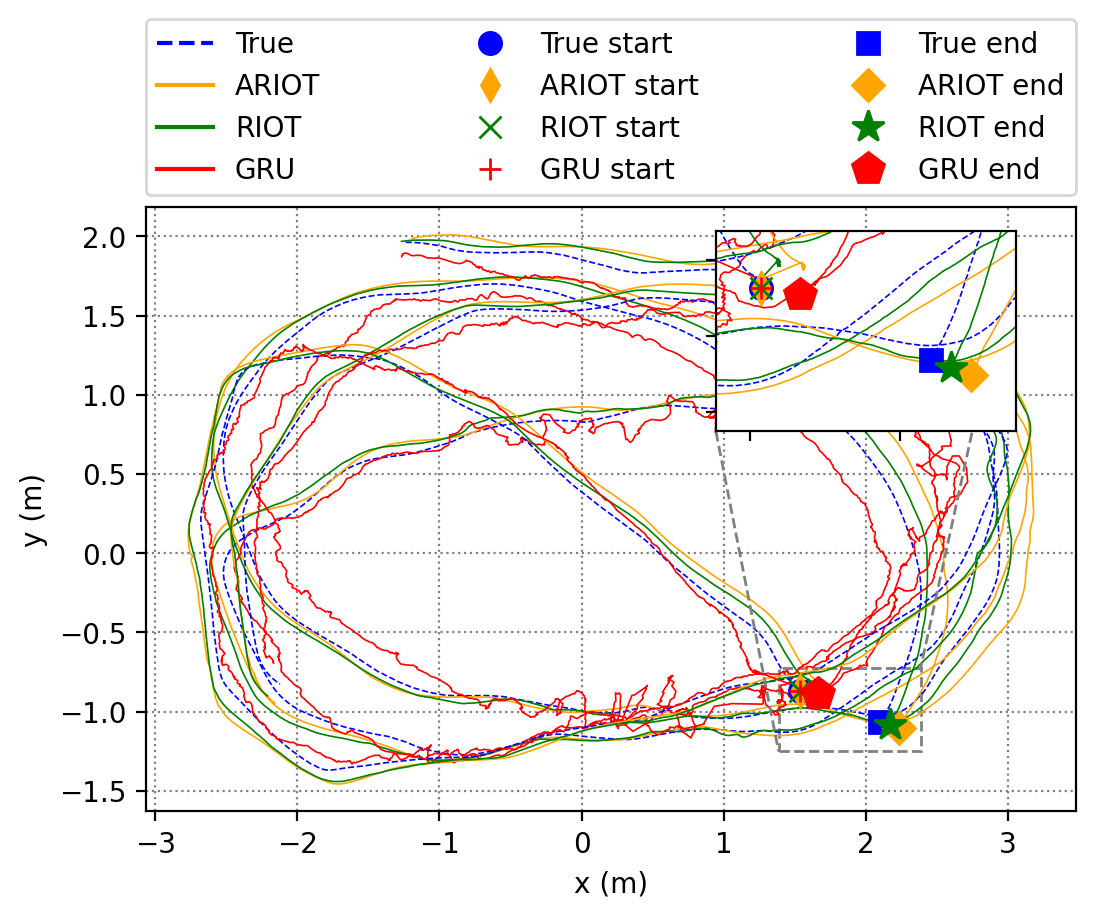}
\caption{Handbag: Total Time: 365.8s, 
TRUE Distance: 305.9m, 
ARIOT Distance: 321.9m, 
RIOT Distance: 317.1m, 
GRU Distance: 341.1m}
\end{minipage}
\end{subfigure}
\hfill
\begin{subfigure}{0.4\textwidth}
\begin{minipage}[t]{\textwidth}
\includegraphics[width=\textwidth]{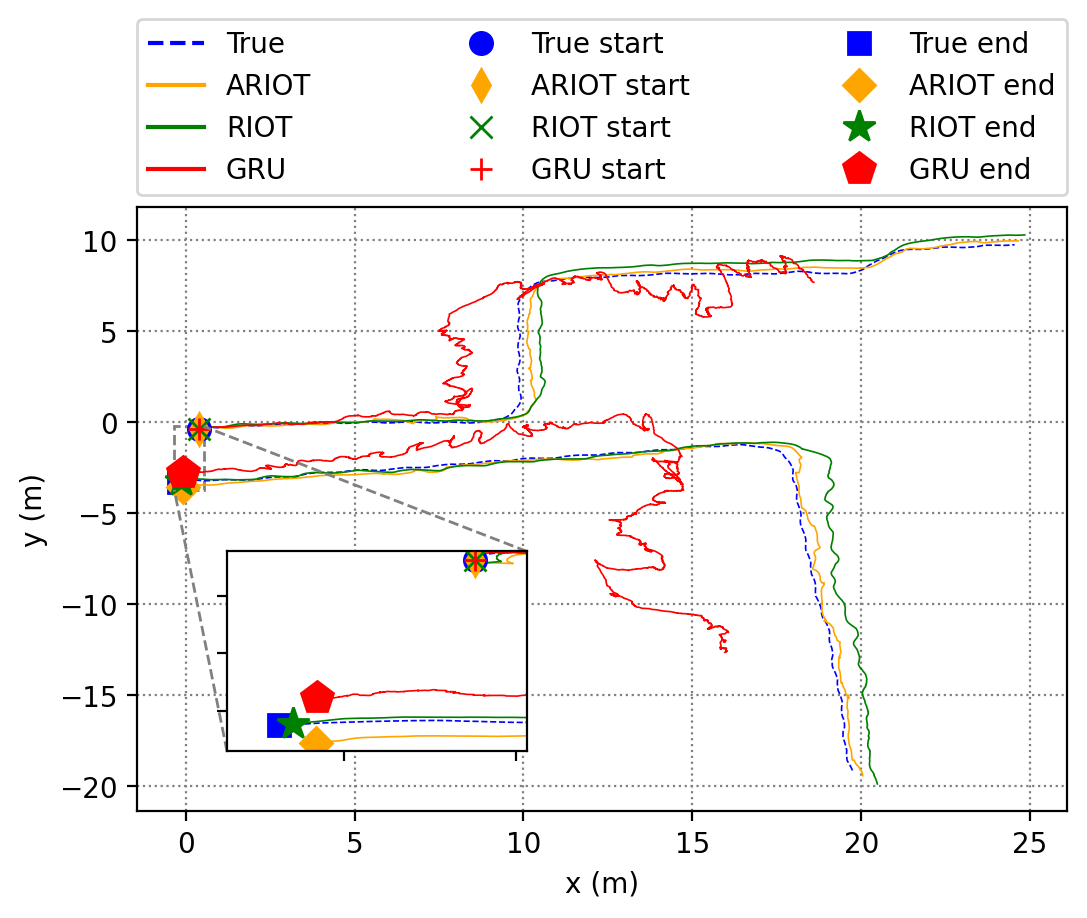}
\caption{Handheld: Total Time: 156.0s, 
TRUE Distance: 152.8m, 
ARIOT Distance: 169.7m, 
RIOT Distance: 159.8m, 
GRU Distance: 358.6m}
\end{minipage}
\end{subfigure}

\vspace{1cm}

\begin{subfigure}{0.4\textwidth}
\begin{minipage}[t]{\textwidth}
\includegraphics[width=\textwidth]{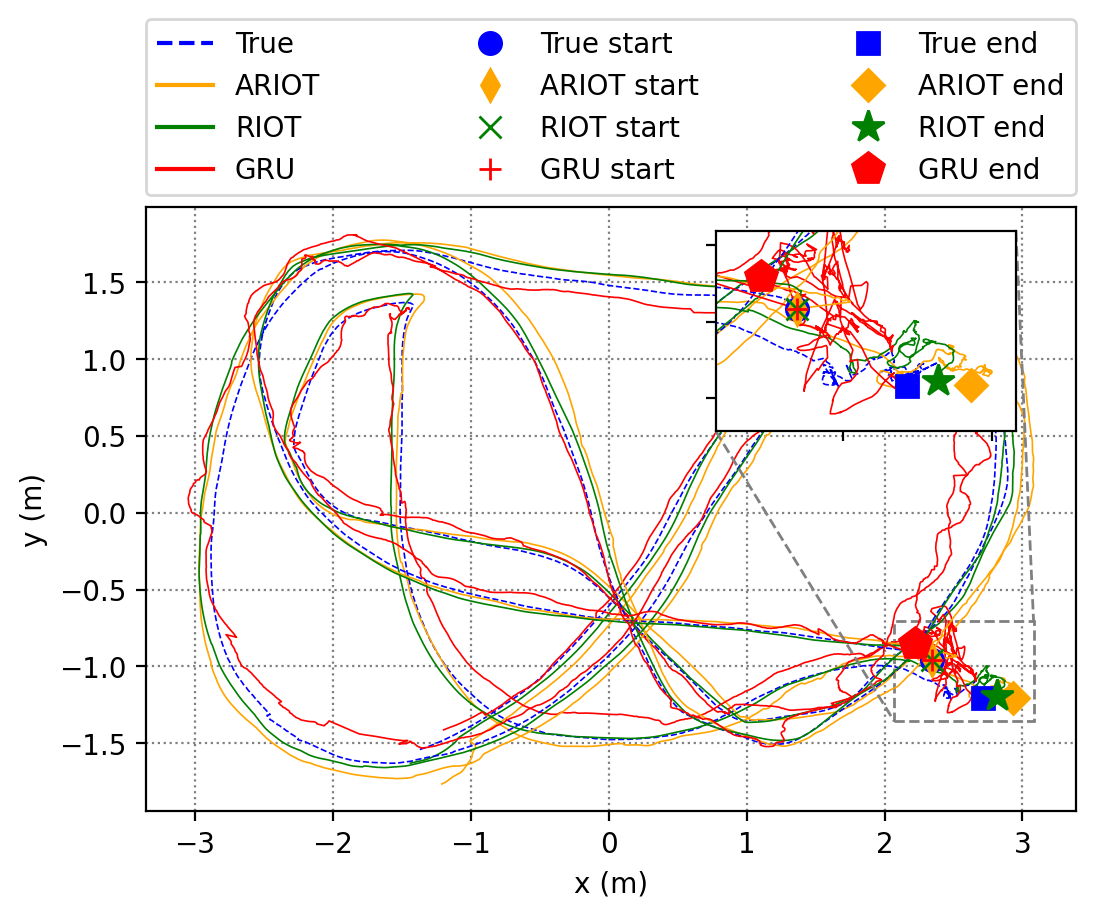}
\caption{iPhone 5: Total Time: 183.8s, 
TRUE Distance: 142.4m, 
ARIOT Distance: 149.8m, 
RIOT Distance: 145.3m, 
GRU Distance: 158.3m} 
\end{minipage}
\end{subfigure}
\hfill
\begin{subfigure}{0.4\textwidth}
\begin{minipage}[t]{\textwidth}
\includegraphics[width=\textwidth]{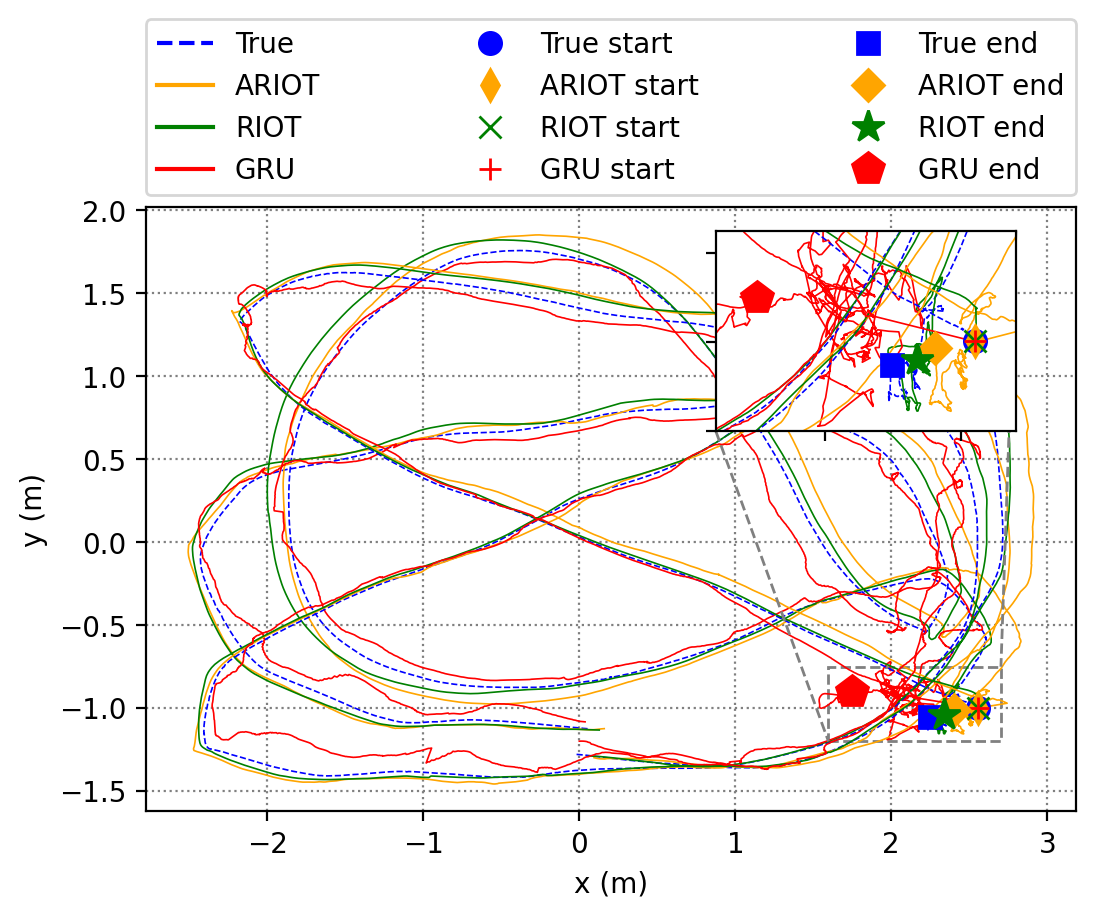}
\caption{iPhone 6: Total Time: 173.4s, 
TRUE Distance: 133.4m, 
ARIOT Distance: 141.6m, 
RIOT Distance: 137.4m, 
GRU Distance: 150.3m}
\end{minipage}
\end{subfigure}

\caption{Position estimates from the first and last minute of each network as well as the true path. The sequence run time is given as well as each approaches total path length.}
\label{fig:figures}
\end{figure*}

\bibliographystyle{unsrtnat}
\bibliography{RIOT}  

\end{document}